\pdfoutput=1
\documentclass[lettersize,journal]{IEEEtran}
\usepackage{amsmath,amsfonts,amssymb}
\usepackage{algorithmic}
\usepackage{algorithm}
\usepackage{array}
\usepackage[caption=false,font=normalsize,labelfont=sf,textfont=sf]{subfig}
\usepackage{textcomp}
\usepackage{stfloats}
\usepackage{url}
\usepackage{verbatim}
\usepackage{graphicx}
\usepackage{booktabs}
\usepackage{cite}
\usepackage{xcolor}
\usepackage{hyperref}
\hypersetup{
    colorlinks=true,
    linkcolor=black,
    filecolor=black,
    urlcolor=black,
    citecolor=black,
}
\usepackage{mathtools}
\usepackage{amsthm}

\newtheorem{myprob}{Problem}

\newtheorem{myrema}{Remark}

\newcommand{\E}{\mathbb{E}}
\newcommand{\R}{\mathbb{R}}
\newcommand{\pos}[1]{\left[#1\right]_{+}}

\begin{document}

\title{Power Setpoint Tracking for Large-Scale EV Charging: Transformer-Constrained Charging Policy Learning with Self-Tuning Constraint Prices}

\author{Daniyal~Sajedi-Hosseini, Yoshihiko Fujisawa,
         Shubham Singh, Lei Zhou, Katsuki Fujisawa, Sebastien Gros, Mazaher Karimi, and Xun~Shen

\thanks{This work has been submitted to the IEEE for possible publication. Copyright may be transferred without notice, after which this version may no longer be accessible.}

\thanks{(Corresponding author: Xun Shen.)}

\thanks{D.~Sajedi-Hosseini and X.~Shen are with the Graduate School of Bio-Applications and Systems Engineering, Tokyo University of Agriculture and Technology, Koganei, Tokyo, Japan (e-mail: s265021u@st.go.tuat.ac.jp).}

\thanks{Y.~Fujisawa, S.~Singh, L.~Zhou, and K.~Fujisawa are with the Institute of Science Tokyo, Yokohama 226-8501, Japan.}

\thanks{S.~Gros is with the Norwegian University of Science and Technology, 7491 Trondheim, Norway.}

\thanks{M.~Karimi is with School of Technology and Innovations, University of Vaasa, Wolffintie 32, 65200 Vaasa, Finland.}

}

\markboth{Preprint submitted to IEEE Transactions on Transportation Electrification}%
{Sajedi-Hosseini \MakeLowercase{\textit{et al.}}: Transformer-Constrained Charging Policy Learning for Power Setpoint Tracking}

\maketitle

\begin{abstract}
In power setpoint tracking (PST), an operator follows a purchased charging profile without overloading shared transformers or undercharging vehicles. PST is cast as a constrained Markov decision process that minimizes tracking error subject to overload and mean user-satisfaction constraints. Transformer-Constrained Charging Policy Learning (TC-CPL) extends soft actor-critic with proportional--integral Lagrange-multiplier updates from measured violations, a residual on the incumbent dispatcher that is warm-started from its demonstrations, and a transformer-aware encoder shared by all ports. It observes no departure times or states of charge. Simulations (five seeds, $100$ paired days) compare it with the incumbent, Round-Robin. On a $600$-port, $122$-substation district, raw overload falls from $3.59$ to $0.0017$~kWh/day (overloaded seed-days from $77\%$ to $0.4\%$), and absolute tracking error falls by $27\%$. Delivered energy rises by $16\%$, and satisfaction rises from $0.959$ to $0.989$. On a $42$-kW transformer with household load, overload falls from $1.86$ to $0.0014$~kWh/day and satisfaction rises from $0.974$ to $0.985$, but absolute tracking error is $22\%$ higher. Against hand-set fixed multipliers, self-tuning is statistically indistinguishable on the district. On the single transformer, it raises satisfaction by $0.011$ at similar overload, but a short sweep found a better-serving fixed pair. The overload multipliers stay below their cap.
\end{abstract}


\section{Introduction}
\label{sec:intro}

\IEEEPARstart{E}{lectric} vehicle (EV) sales passed $20$ million in $2025$~\cite{iea2026}. EV charging raises peak loading and shortens transformer life~\cite{li2022transformer}. Public chargers share transformers with households, so the charging headroom (each rating minus the inflexible load) varies over the day and shrinks during demand response. Only dedicated depots have their own transformers, and their demand follows planned routes~\cite{shojaeefard2022tsp}. The operator buys electricity a day ahead and then follows the committed profile, a task known as power setpoint tracking (PST).

\subsection{Related Work}
\label{ssec:related}

Rule-based charging, such as charge-as-fast-as-possible (CAFAP) and Round-Robin~\cite{ev2gym}, is cheap but ignores transformer limits. Time--cost cloud-workflow schedulers are as fast but need all task durations in advance, which charging sessions do not reveal~\cite{mollajafari2021tc3pop}.
Model-based methods include model predictive control (MPC)~\cite{tang2017,diazlondono2026} and distributionally robust MPC~\cite{nguyen2023distributionally}. They state constraints explicitly but need departure times and energy requirements that public-charging operators rarely know. Their online optimization also grows with the number of chargers.

Once trained, model-free reinforcement learning (RL) acts with one forward pass, which suits real-time operation~\cite{sutton2018}. RL has been applied to EV charging scheduling~\cite{wan2019,jin2021}, coordination~\cite{sadeghianpourhamami2020} and management~\cite{li2022drl}. Y{\i}lmaz et al.~\cite{yilmaz2024} applied deep deterministic policy gradient to PST in EV2Gym (DDPG-PST).
Table~\ref{tab:positioning} also lists recent multi-agent, distributed, federated, robust and physics-informed learners. Most fix the grid and user reward weights before training, as does a battery-aware reward for hybrid-storage EV energy management~\cite{raeesi2026beyond}. None targets setpoint tracking, so the learned baselines are DDPG-PST, proximal policy optimization (PPO-PST) and constrained policy optimization (CPO)~\cite{achiam2017}.

Constrained Markov decision processes (CMDPs)~\cite{altman1999} separate what a policy should optimize from what it must not violate. Surveys cover safe RL in general~\cite{garcia2015} and in power systems~\cite{su2025review}. CPO keeps the policy near-feasible at every trust-region update. Tessler et al.~\cite{tessler2019} learn the penalty weight on a slower timescale rather than by trial and error or offline search, such as genetic-algorithm tuning~\cite{nazemi2022ga}. Stooke et al.~\cite{stooke2020} showed that this dual update is integral-only feedback and added proportional and derivative terms to damp oscillation.

\begin{table*}[!t]
\centering
\caption{RL charging methods vs.\ TC-CPL (DDPG-PST as run here). Real-time = online per interval; Dep., SoC = departure time, state of charge as input; \checkmark/$\times$ = yes/no; n/r = not reported.}
\label{tab:positioning}
\footnotesize
\setlength{\tabcolsep}{1pt}
\begin{tabular*}{\textwidth}{@{\extracolsep{\fill}}llclccllcc@{}}
\toprule
Reference & Algorithm & Real-time & Constraint handling & Dep. & SoC & Transf.\ limit & Warm start & V2G & Open-source \\
\midrule
\cite{li2020safe} & CPO & \checkmark & full charge at departure & $\times$ & \checkmark & $\times$ & n/r & \checkmark & $\times$ \\
\cite{zhang2023safe} & constrained SAC & \checkmark & Lagrangian + safety filter & n/r & n/r & n/r & n/r & \checkmark & $\times$ \\
\cite{chen2025augmented} & SAC & \checkmark & aug.\ Lagrangian, dual step per update & $\times$ & \checkmark & $\times$ & n/r & \checkmark & $\times$ \\
\cite{fan2024safety} & SAC & \checkmark & bus-voltage Lagrangian & \checkmark & \checkmark & $\times$ & n/r & \checkmark & $\times$ \\
\cite{yan2022cooperative} & multi-agent SAC & \checkmark & overload penalty & n/r & n/r & \checkmark{} penalty & n/r & n/r & $\times$ \\
\cite{li2022transformer} & multi-agent TD3 & \checkmark & fixed-weight aging + user costs & \checkmark & \checkmark & \checkmark{} aging & curriculum & \checkmark & $\times$ \\
\cite{ye2022cade} & DQN & \checkmark & unmet-energy penalty & \checkmark & \checkmark & $\times$ & $\times$ & \checkmark & $\times$ \\
\cite{zhang2023stackelberg} & Stackelberg multi-agent & \checkmark & ramping penalty & n/r & n/r & n/r & n/r & n/r & $\times$ \\
\cite{qian2024federated} & federated SAC & \checkmark & OPF + reward penalties & \checkmark & \checkmark & $\times$ & n/r & \checkmark & $\times$ \\
\cite{li2025robust} & robust multi-agent & \checkmark & bi-level price signals & n/r & n/r & n/r & n/r & n/r & $\times$ \\
\cite{ORFANOUDAKIS2026128224} & physics-informed TD3 & \checkmark & fixed-weight voltage penalty & \checkmark & \checkmark & $\times$ & n/r & \checkmark & \checkmark \\
\cite{yilmaz2024} & DDPG & \checkmark & none (tracking reward) & $\times$ & $\times$ & $\times$ & $\times$ & $\times$ & \checkmark \\
TC-CPL (ours) & SAC & \checkmark & PI multipliers, measured violations & $\times$ & $\times$ & \checkmark{} time-varying & \checkmark{} demos + residual & $\times$ & \checkmark \\
\bottomrule
\end{tabular*}
\end{table*}

\label{ssec:positioning}%
Table~\ref{tab:positioning} compares TC-CPL with RL charging methods. Two of them penalize transformer overload~\cite{yan2022cooperative} or aging~\cite{li2022transformer} in the reward. None treats the transformer limit as a constraint. The closest, Chen et al.~\cite{chen2025augmented}, controls one EV with an augmented-Lagrangian soft actor-critic (SAC) that takes a dual step at every network update. TC-CPL instead updates the multipliers once per window of episodes with a proportional--integral (PI) controller driven by measured violations. It also learns a residual on the site's dispatcher, warm-started from its demonstrations, and shares one transformer-aware encoder across all ports and transformers. These building blocks are known. The contribution is their formulation for transformer-constrained PST under partial information, the two-level port--transformer encoder, and the attainability condition this combination places on the PI multiplier.

\subsection{Proposed Approach and Contributions}
\label{ssec:contrib}

TC-CPL addresses safety and deployment challenges of RL that remain open in EV dispatch~\cite{qiu2023}:
\begin{itemize}
    \item \textbf{Transformer-constrained PST with self-tuned multipliers.} PST is posed as a CMDP that uses site measurements and a non-EV load nowcast but no departure times or states of charge. A PI controller tunes the two multipliers with one setting for both scenarios. A fixed pair can do as well, but only after a scenario-specific sweep.

    \item \textbf{A residual, transformer-aware architecture with a port-independent parameter count.} A zero residual reproduces the incumbent, but neither the incumbent nor its demonstrations make exploration safe. With shared weights, the actor has $117{,}634$ parameters for $20$ or $600$ ports. A full $600$-port dispatch takes $2.3$~ms (upper bound) on one laptop CPU thread.

    \item \textbf{Validation in two transformer-constrained scenarios.} The scenarios are a $42$-kW transformer with household load, photovoltaics and demand response (\texttt{PublicPST-S}) and a $122$-substation, $600$-port district (\texttt{cityNL}). TC-CPL is compared with five baselines and the offline optimum over five seeds on the same $100$ days, with paired confidence intervals. The paper also reports computational cost. On the single transformer, it tests robustness and multiplier sensitivity, and its ablations show that removing residual anchoring or the load-aware inputs improves service.
\end{itemize}

\section{Formulation of Transformer-Constrained PST}
\label{sec:problem}

\begin{figure}[!t]
\centering
\includegraphics[width=\columnwidth]{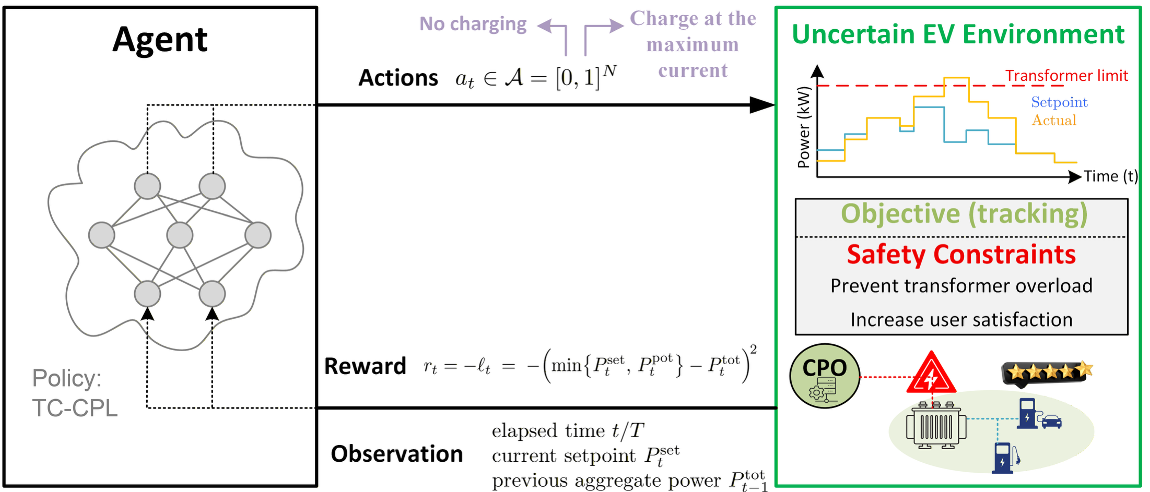}
\caption{Transformer-constrained PST. The observation box shows part of $o_t$. ``CPO'' is the charge point operator, not the baseline.}
\label{fig:Problem_statement}
\end{figure}

\begin{figure*}[!t]
\centering
\includegraphics[width=\textwidth]{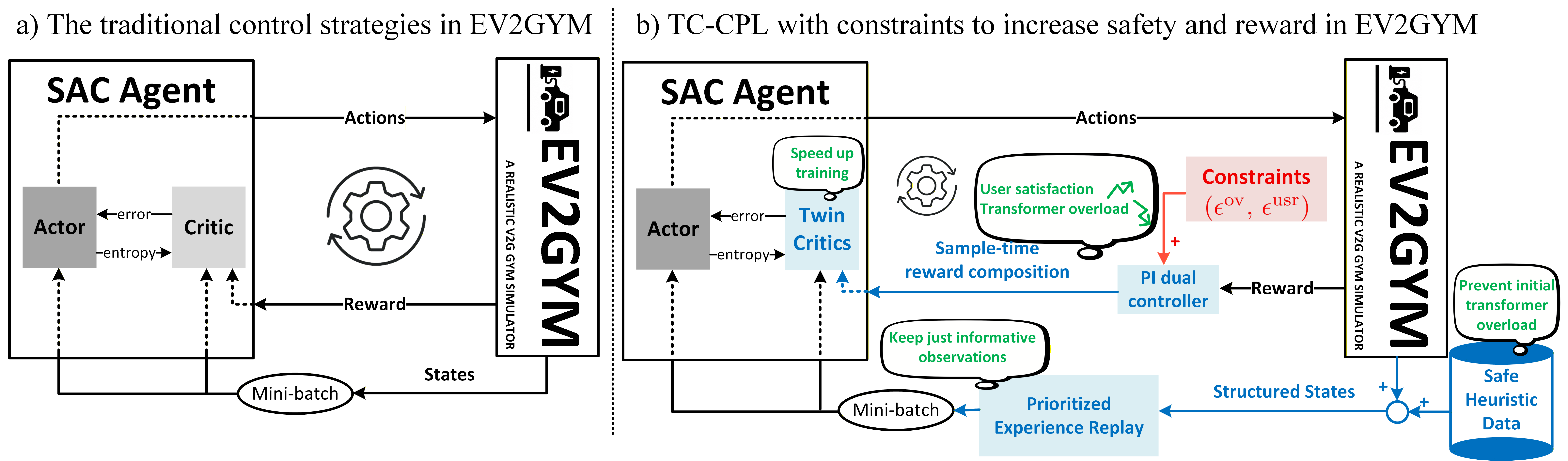}
\caption{(a) Conventional SAC with fixed penalties. (b) TC-CPL adds PI-controlled multipliers (M1), incumbent demonstrations (M2, labeled ``safe'' but not overload-free) and grid-aware states (M3).}
\label{fig:comparison_methods}
\end{figure*}

The operator dispatches charging power to track its purchased setpoint $\{P^{\mathrm{set}}_t\}_{t \in \mathcal{T}}$ despite uncertain arrivals, departures and demands (Fig.~\ref{fig:Problem_statement}). Tracking is the objective, and transformer loading and satisfaction are constraints. Conventional RL folds such constraints into fixed reward penalties (Fig.~\ref{fig:comparison_methods}(a))~\cite{ORFANOUDAKIS2026128224}.

\subsection{Transformer-Constrained PST as a CMDP}
\label{ssec:statement}

In this episodic CMDP, a step is a dispatch interval of length $\Delta t$, and an episode is a day of $T$ intervals, $t\in\mathcal{T}=\{0,\dots,T-1\}$. EV supply equipment (EVSE) ports are indexed by $j \in \mathcal{J}$ ($N=|\mathcal{J}|$), transformers by $w \in \mathcal{W}$ and EVs by $k \in \mathcal{E}$.

\subsubsection{\textbf{State space}}
The state has action-dependent and exogenous parts,
\begin{align}
    s_t &= (x_t,\, \xi_t) \in \mathcal{S},
    \label{eq:state-split}\\
    x_t &= \bigl[\,\{E_{j,t}\}_{j \in \mathcal{J}},\;
    \{u_{j,t}\}_{j \in \mathcal{J}}\,\bigr],
    \label{eq:state-x}\\
    \xi_t &= \bigl[\, t,\; \{P^{\mathrm{set}}_{t'}\}_{t' \ge t},\;
    \{t^{\mathrm{arr}}_k,\, t^{\mathrm{dep}}_k,\, E^{\mathrm{arr}}_k,\,\overline{E}_k\}_{k \in \mathcal{E}}, \notag\\
    &\qquad\;\; \{P^{L}_{w,t},\, P^{PV}_{w,t},\,
    \overline{P}^{\mathrm{tr}}_{w,t}\}_{w \in \mathcal{W}} \,\bigr],
    \label{eq:state-xi}
\end{align}
where $E_{j,t}$ and $u_{j,t}$ are the battery energy and occupancy of EVSE $j$; $t^{\mathrm{arr}}_k$, $t^{\mathrm{dep}}_k$, $E^{\mathrm{arr}}_k$ and $\overline{E}_k$ the arrival and departure times, arrival energy and capacity of EV $k$; and $P^{L}_{w,t}$, $P^{PV}_{w,t}$ and $\overline{P}^{\mathrm{tr}}_{w,t}$ the inflexible load, photovoltaic (PV) generation and limit of transformer $w$. With $P^{PV}_{w,t}\ge0$, the non-EV load is $P^{\mathrm{nEV}}_{w,t}=P^{L}_{w,t}-P^{PV}_{w,t}$. The limit varies in time and drops during demand-response (DR) events (Table~\ref{tab:setup}).

\subsubsection{\textbf{Observation space}}
The observation $o_t$ (Section~\ref{ssec:m3}) extends the EV2Gym PST state of DDPG-PST~\cite{yilmaz2024}, which holds $t/T$, $P^{\mathrm{set}}_t$, $P^{\mathrm{tot}}_{t-1}$ and, per port, the charger status and the energy and time since plug-in. It adds the charging potential $P^{\mathrm{pot}}_t$, which needs each vehicle's AC rating, and, per transformer, the limit in force, its announced minimum over a look-ahead window and the non-EV load. The policy does not observe $E^{\mathrm{arr}}_k$, $\overline{E}_k$, $t^{\mathrm{dep}}_k$, the state of charge beyond the charger status or future arrivals. The departure state of charge enters only the training costs and the metrics. As in EV2Gym, $P^{\mathrm{pot}}_t$ omits vehicles in their last connected interval, a one-interval look-ahead on departures.
Admissible policies are causal,
\begin{equation}
    \label{eq:policyclass}
    \Pi = \bigl\{\, \pi : (o_0,\dots,o_t) \mapsto a_t \,\bigr\}.
\end{equation}

\subsubsection{\textbf{Action space}}
An action $a_t \in \mathcal{A} = [0,1]^N$ sets each EVSE's fraction of its admissible charging current (Section~\ref{ssec:pst}). Vehicle-to-grid (V2G) is disabled. The hardware enforces the EVSE and station current limits but not the transformer limit.

\subsubsection{\textbf{Transition dynamics}}
The state evolves as
\begin{align}
    \label{eq:transition}
    x_{t+1} &= f(x_t, \xi_t, a_t), \qquad \xi_{t+1} \sim \mathcal{D}(\cdot \mid \xi_t),\\
    \label{eq:transition_kernel}
    \mathcal{P}(s_{t+1} \mid s_t, a_t) &=
    \delta_{\mathrm{D}}\bigl(x_{t+1} - f(x_t,\xi_t,a_t)\bigr)\, \mathcal{D}(\xi_{t+1} \mid \xi_t),
\end{align}
with the Dirac delta $\delta_{\mathrm{D}}$, the charging dynamics $f$~\eqref{eq:current}--\eqref{eq:soc} and an unknown exogenous law $\mathcal{D}$.

\subsubsection{\textbf{Tracking objective}}
The objective is the negative squared setpoint error, as in~\cite{yilmaz2024},
\begin{equation}
    \label{eq:stagecost}
    r_t = -\ell_t  \;=\; -\Bigl(\min\bigl\{P^{\mathrm{set}}_t,\, P^{\mathrm{pot}}_t\bigr\} - P^{\mathrm{tot}}_t\Bigr)^{\!2},
\end{equation}
where $P^{\mathrm{tot}}_t$ is the aggregate charging power and $P^{\mathrm{pot}}_t$ the most the connected fleet could absorb~\eqref{eq:ptot}. The reported absolute and squared errors use the setpoint itself: $|\epsilon_{\mathrm{tr}}|=\Delta t\sum_{t}|P^{\mathrm{set}}_t-P^{\mathrm{tot}}_t|$ (kWh/day) and $\epsilon_{\mathrm{tr}}=\sum_{t}(P^{\mathrm{set}}_t-P^{\mathrm{tot}}_t)^2$ (kW$^2$).

\subsubsection{\textbf{Operational constraints}}
The episode overload energy is
\begin{equation}
    \label{eq:overload}
    \epsilon^{\mathrm{ov}} \;=\; \sum_{t \in \mathcal{T}} \sum_{w \in \mathcal{W}} \pos{\,P^{\mathrm{tr}}_{w,t} - \overline{P}^{\mathrm{tr}}_{w,t}}\,\Delta t \qquad \text{[kWh]},
\end{equation}
with $\pos{x} \triangleq \max\{x,0\}$ and the loading $P^{\mathrm{tr}}_{w,t}$, the EV power plus $P^{\mathrm{nEV}}_{w,t}$~\eqref{eq:trload}.
With $\epsilon^{\mathrm{ov}}_{\max}=0$, the expected-overload constraint~\eqref{eq:cmdp-c1} demands zero overload almost surely. A learned policy meets it only approximately, with no per-interval guarantee. Section~\ref{ssec:setup} also reports per-interval excesses, and Section~\ref{ssec:robust} tests an optional projection that enforces the limit at dispatch.

The transformer is represented only by this time-varying limit. Hot-spot temperature, overload duration, ambient conditions and aging are not modeled, so $\epsilon^{\mathrm{ov}}$ is only a proxy for thermal stress. Loading guides permit short, moderate overloads, depending on the unit's thermal state~\cite{iec600767}.
Service is measured by the departure state of charge~\cite{ev2gym} relative to the expected level $\mathrm{SoC}^{\star}_k$,
\begin{equation}
    \label{eq:usr}
    s_k = \min\Bigl\{ \tfrac{\mathrm{SoC}_{k,\,t_k^{\mathrm{dep}}}}{\mathrm{SoC}^{\star}_k},\, 1\Bigr\}, \quad
    \epsilon^{\mathrm{usr}} = \frac{1}{|\mathcal{E}|} \sum_{k \in \mathcal{E}} s_k.
\end{equation}
A fleet average avoids per-vehicle constraints with unobserved departure times. Because it can hide poorly served vehicles, Section~\ref{ssec:setup} also reports per-vehicle service metrics. In upper-bound form,
\begin{equation}
\label{eq:deficit}
\epsilon^{\mathrm{def}} \triangleq 1-\epsilon^{\mathrm{usr}},
\qquad
\epsilon^{\mathrm{def}}_{\max} \triangleq 1-\epsilon^{\mathrm{usr}}_{\min}.
\end{equation}

\begin{myprob}[\textbf{Transformer-Constrained PST}]
\label{prob:TC-CPL}
Find
\begin{subequations}
\label{eq:cmdp}
\begin{align}
\pi^{\star} \in \arg\max_{\pi \in \Pi}\;\;
& J(\pi) = \E_{\tau \sim \pi}\!\Bigl[\,\textstyle\sum_{t \in \mathcal{T}} r_t\Bigr]
\label{eq:cmdp-obj}\\
\text{s.t.}\;\;
& G_1(\pi) = \E_{\tau \sim \pi}\!\bigl[\epsilon^{\mathrm{ov}}\bigr] \;\le\; \epsilon^{\mathrm{ov}}_{\max},
\label{eq:cmdp-c1}\\
& G_2(\pi) = \E_{\tau \sim \pi}\!\bigl[\epsilon^{\mathrm{def}}\bigr] \;\le\; \epsilon^{\mathrm{def}}_{\max},
\label{eq:cmdp-c2}
\end{align}
\end{subequations}
with the operating requirement of zero overload, $\epsilon^{\mathrm{ov}}_{\max}=0$, and a satisfaction floor $\epsilon^{\mathrm{usr}}_{\min}$ (Table~\ref{tab:setup}).
\end{myprob}

\begin{myrema}[Surrogate and feasibility]
\label{rem:slater}
Training uses a surrogate (Section~\ref{ssec:lagrangian}). It sets the overload tolerance to $\tau^{\mathrm{ov}}=0$ above the tightened limit $\widetilde{P}^{\mathrm{tr}}_{w,t}=\max\{(1-m)\overline{P}^{\mathrm{tr}}_{w,t},P^{\mathrm{nEV}}_{w,t}\}$, with back-off $m$, and sets a training floor $\epsilon^{\mathrm{usr}}_{\mathrm{train}}$ above $\epsilon^{\mathrm{usr}}_{\min}$ (Table~\ref{tab:setup}).
(i)~Overload is nonnegative pathwise, so $\tau^{\mathrm{ov}}=0$ admits no strictly feasible point (Slater's condition fails). The multiplier $\lambda_{\mathrm{ov}}$ thus need not converge to a finite value and is not read as a shadow price (Section~\ref{ssec:adaptation}).
(ii)~The PI update can settle only if the surrogate's unavoidable overload, the infimum of its expected overload over the policy class, is zero (Section~\ref{ssec:m1}). This holds for a residual radius $\rho\ge1$~\eqref{eq:projection}, whose class contains zero charging. For a smaller radius, it can fail, and $\lambda_{\mathrm{ov}}$ then rises toward its cap. This rise signals an unattainable target, not a binding constraint (Sections~\ref{ssec:adaptation} and~\ref{ssec:ablation}).
(iii)~Zero overload above $\widetilde{P}^{\mathrm{tr}}_{w,t}$ meets the raw limit strictly, with slack $m\overline{P}^{\mathrm{tr}}_{w,t}$ or more, wherever $P^{\mathrm{nEV}}_{w,t}<(1-m)\overline{P}^{\mathrm{tr}}_{w,t}$.
(iv)~Joint feasibility of both constraints for the causal class~\eqref{eq:policyclass} is not proven. Section~\ref{ssec:setup} checks it in hindsight for the raw limit.
\end{myrema}

Because $\mathcal{D}$ is unknown, $J$, $G_1$ and $G_2$ are estimated from simulated trajectories.

\subsection{Challenges of Transformer-Constrained PST}
\label{ssec:challenges}

Problem~\ref{prob:TC-CPL} poses four difficulties. Table~\ref{tab:positioning} maps RL charging methods to the first three.

\subsubsection{Partial information and uncertainty}
With future arrivals, departure times and states of charge unobserved, forecast-based MPC~\cite{nguyen2023distributionally} lacks inputs, and the hindsight problem~\eqref{eq:offline} serves only as a benchmark.

\subsubsection{Adaptive constraint handling}
How much tracking should yield to loading and service depends on operating conditions and is hard to set beforehand~\cite{qiu2023}. Each Lagrange multiplier should therefore follow measured performance.

\subsubsection{Cold-start operation}
No hardware stops a poor early action from overloading a transformer. A learner can start from the site dispatcher's records, which reflect established, though not overload-free, operation.

\subsubsection{Large-scale charging systems}
Input and output dimensions grow with the number of EVSEs. A scalable network should exploit EVSE interchangeability and keep transformer-level information.

\section{Proposed Method}
\label{sec:method}

Three mechanisms surround an off-policy, model-free soft actor-critic (Fig.~\ref{fig:comparison_methods}(b)). M1 sets the two Lagrange multipliers with a fixed-gain PI controller driven by violations measured in training (Challenge~2). M2 seeds the replay buffer with the site dispatcher's transitions and anchors the action at its command, so exploration is centered on established behavior (Challenge~3). M3 adds the transformer limits and non-EV loads to the observation (Challenge~1) and encodes the ports with shared weights and order-invariant pooling, so the parameter count does not depend on the number of ports (Challenge~4).

\subsection{Power Setpoint Tracking}
\label{ssec:pst}

Charging station $i\in\mathcal{C}$ hosts the EVSEs $\mathcal{J}_i\subseteq\mathcal{J}$, and transformer $w$ feeds the EVSEs $\mathcal{J}_w$. EV $k=k(j,t)$ is connected to EVSE $j$ at step $t$ when $u_{j,t}=1$.

A station shares its current limit $\overline{I}^{\mathrm{cs}}_i$ among its EVSEs, so the action commands, for $j\in\mathcal{J}_i$,
\begin{equation}
\label{eq:current}
\begin{aligned}
I^{\mathrm{cmd}}_{j,t}&=\frac{\overline{I}^{\mathrm{cs}}_i\,a_{j,t}\,u_{j,t}}{\max\bigl\{1,\sum_{j'\in\mathcal{J}_i}a_{j',t}u_{j',t}\bigr\}},\\
0&\le I^{\mathrm{ch}}_{j,t}\le I^{\mathrm{cmd}}_{j,t},
\end{aligned}
\end{equation}
where the vehicle draws less when its AC rating $\overline{P}^{\mathrm{EV}}_k$ binds or when EV2Gym's constant-voltage taper acts at a high state of charge~\cite{ev2gym}. The grid-side and battery-side powers are
\begin{equation}
\label{eq:chpower}
P^{\mathrm{grid}}_{j,t}=V_j\sqrt{\phi_j}\,I^{\mathrm{ch}}_{j,t}\,u_{j,t},
\qquad
P^{\mathrm{ch}}_{j,t}=\eta_k\,P^{\mathrm{grid}}_{j,t},
\end{equation}
with line-to-line voltage $V_j$, $\phi_j$ phases and efficiency $\eta_k\le1$. The station rating is $\overline{P}^{\mathrm{cs}}_i=V_j\sqrt{\phi_j}\,\overline{I}^{\mathrm{cs}}_i$ (values in Table~\ref{tab:setup}).
The battery energy evolves with the battery-side power,
\begin{equation}
\label{eq:soc}
E_{j,t+1} \;=\; E_{j,t} + P^{\mathrm{ch}}_{j,t}\,\Delta t, \qquad E_{j,\,t^{\mathrm{arr}}_k} = E^{\mathrm{arr}}_k,
\end{equation}
with $E_{j,t}\le\overline{E}_k$, and $\mathrm{SoC}_{k,t} = E_{j,t}/\overline{E}_k$. The aggregate power and the charging potential of Section~\ref{ssec:statement} are
\begin{equation}
\label{eq:ptot}
\begin{aligned}
P^{\mathrm{tot}}_t &= \textstyle\sum_{j \in \mathcal{J}} P^{\mathrm{ch}}_{j,t},\\
P^{\mathrm{pot}}_t &= \textstyle\sum_{i\in\mathcal{C}}\min\Bigl\{\overline{P}^{\mathrm{cs}}_i,\,
\sum_{j\in\mathcal{J}_i}\bar u_{j,t}\min\bigl\{\overline{P}^{\mathrm{cs}}_i,\overline{P}^{\mathrm{EV}}_k\bigr\}\Bigr\},
\end{aligned}
\end{equation}
where $\bar u_{j,t}\in\{0,1\}$ marks a vehicle with $\mathrm{SoC}_{k,t}<1$ and $t^{\mathrm{dep}}_k>t$. Unlike $P^{\mathrm{tot}}_t$, the potential $P^{\mathrm{pot}}_t$ is grid-side.

A transformer physically carries the grid-side power, but EV2Gym, used as in~\cite{yilmaz2024}, books the battery-side $P^{\mathrm{ch}}_{j,t}$ in $P^{\mathrm{tot}}_t$ and in the transformer loading~\eqref{eq:trload}. The difference is at most a fraction $1/\eta_k-1$ of the EV load. Where $P^{\mathrm{nEV}}_{w,t}\ge0$ and $\eta_k\ge1-m$, the back-off $m$ of the tightened limit (Remark~\ref{rem:slater}) covers these losses, since $P^{\mathrm{tr}}_{w,t}\le(1-m)\overline{P}^{\mathrm{tr}}_{w,t}$ implies
\begin{equation}
\label{eq:3-gridbound}
P^{\mathrm{nEV}}_{w,t}+\textstyle\sum_{j\in\mathcal{J}_w}P^{\mathrm{ch}}_{j,t}/\eta_k
\le P^{\mathrm{tr}}_{w,t}/(1-m)
\le\overline{P}^{\mathrm{tr}}_{w,t}.
\end{equation}
Section~\ref{ssec:publicpst} also reports the overload recounted grid-side.

With the day fully known, the delivered power $p=\{P^{\mathrm{ch}}_{j,t}\}$ is scheduled in hindsight by
\begin{subequations}
\label{eq:offline}
\begin{align}
J^{\star}_{0}=\min_{p}\;&\textstyle\sum_{t\in\mathcal{T}}\bigl(P^{\mathrm{set}}_t-\sum_{j\in\mathcal{J}}P^{\mathrm{ch}}_{j,t}\bigr)^{2}
\notag\\
\text{s.t.}\;\;&0\le P^{\mathrm{ch}}_{j,t}\le\eta_k\min\bigl\{\overline{P}^{\mathrm{EV}}_k,\overline{P}^{\mathrm{cs}}_i\bigr\}\,u_{j,t},
\label{eq:3-off-port}\\
&\textstyle\sum_{j\in\mathcal{J}_i}P^{\mathrm{ch}}_{j,t}/\eta_k\le\overline{P}^{\mathrm{cs}}_i,
\label{eq:3-off-station}\\
&E^{\mathrm{arr}}_k+\Delta t\textstyle\sum_{t\in\mathcal{T}_k}P^{\mathrm{ch}}_{j_k,t}\le\overline{E}_k,
\label{eq:3-off-battery}\\
&\textstyle\sum_{j\in\mathcal{J}_w}P^{\mathrm{ch}}_{j,t}\le\pos{\overline{P}^{\mathrm{tr}}_{w,t}-P^{\mathrm{nEV}}_{w,t}},
\label{eq:3-off-tr}
\end{align}
\end{subequations}
for all indices, where $j_k$ is the EVSE of EV $k$ and $\mathcal{T}_k$ its connected intervals. Constraint~\eqref{eq:3-off-tr} is a hard transformer limit, so the benchmark never trades overload for service. With V2G disabled,~\eqref{eq:offline} is an exact convex quadratic program (QP) without binary variables. The QP has no satisfaction constraint, and adding an attainable floor changes $J^{\star}_0$ negligibly (Section~\ref{ssec:setup}). The schedule is replayed through EV2Gym and scored like the learned controllers. As constraint~\eqref{eq:3-off-battery} ignores the charging taper, the replayed squared error can exceed $J^{\star}_0$, so the replayed schedule is an attainable reference, not a certified bound.

\subsection{Lagrangian Learning Architecture}
\label{ssec:lagrangian}

Problem~\ref{prob:TC-CPL} is relaxed with multipliers $\lambda = (\lambda_{\mathrm{ov}}, \lambda_{\mathrm{def}}) \succeq 0$~\cite{altman1999}. With the aggregate station rating $P_{\mathrm{ref}}=\sum_{i\in\mathcal{C}}\overline{P}^{\mathrm{cs}}_i$ and $E_{\mathrm{ref}}=T\Delta t P_{\mathrm{ref}}$ (Table~\ref{tab:setup}), let $\bar r_t=\Delta t\,r_t/P_{\mathrm{ref}}^2$ and
\begin{equation}
\label{eq:exact-costs}
\begin{aligned}
c^{\mathrm{ov}}_t
&=\frac{\Delta t}{E_{\mathrm{ref}}}
\sum_{w\in\mathcal{W}}
\pos{P^{\mathrm{tr}}_{w,t}-\widetilde P^{\mathrm{tr}}_{w,t}},\\
c^{\mathrm{def}}_t
&=\mathbb{I}\{t=T-1\}\bigl(1-\epsilon^{\mathrm{usr}}\bigr),
\end{aligned}
\end{equation}
with the tightened limit $\widetilde P^{\mathrm{tr}}_{w,t}$ of Remark~\ref{rem:slater}. As this limit is never below $P^{\mathrm{nEV}}_{w,t}$, only overload that charging can remove is priced. For an episode with at least one departure,
\begin{equation}
\label{eq:exact-cost-identities}
\sum_t c^{\mathrm{ov}}_t=\frac{\tilde\epsilon^{\mathrm{ov}}}{E_{\mathrm{ref}}},
\qquad
\sum_t c^{\mathrm{def}}_t=1-\epsilon^{\mathrm{usr}},
\end{equation}
where $\tilde\epsilon^{\mathrm{ov}}$ is~\eqref{eq:overload} counted above $\widetilde P^{\mathrm{tr}}_{w,t}$. An episode without departures has zero service cost. The thresholds are $d^{\mathrm{ov}}=\tau^{\mathrm{ov}}/E_{\mathrm{ref}}=0$ and $d^{\mathrm{def}}=1-\epsilon^{\mathrm{usr}}_{\mathrm{train}}$, and the factor $T\Delta t$, the length of a day, expresses $\lambda_{\mathrm{ov}}$ per kWh of daily overload in the normalized Lagrangian
\begin{multline}
\label{eq:lagrangian}
\overline{\mathcal{L}}(\pi,\lambda)
=\overline J(\pi)
-\lambda_{\mathrm{ov}}T\Delta t\bigl(\overline G_{\mathrm{ov}}(\pi)-d^{\mathrm{ov}}\bigr)\\
-\lambda_{\mathrm{def}}\bigl(\overline G_{\mathrm{def}}(\pi)-d^{\mathrm{def}}\bigr),
\end{multline}
where $\overline J(\pi)=\E[\sum_t\bar r_t]$ and $\overline G_q(\pi)=\E[\sum_t c^q_t]$.
The terminal form of $c^{\mathrm{def}}_t$ is kept because it is exactly the deficit that~\eqref{eq:cmdp-c2} bounds. Without discounting ($\gamma$ in Table~\ref{tab:setup}) and with the terminal mask of~\eqref{eq:sac-target}, the critics carry it to every step. A shaping term also charges each vehicle $k\in\mathcal{E}^{\mathrm{dep}}_t$ departing at $t$ its shortfall,
\begin{equation}
\label{eq:3-hinge}
c^{\mathrm{aux}}_t=\frac{1}{N}\sum_{k\in\mathcal{E}^{\mathrm{dep}}_t}\pos{\epsilon^{\mathrm{usr}}_{\mathrm{train}}-s_k},
\end{equation}
with weight $\zeta\kappa$, where the factor $\kappa$ falls linearly to zero over the first part of training (Table~\ref{tab:setup}). It carries no multiplier, so it leaves Problem~\ref{prob:TC-CPL} unchanged. No overload shaping is used.

\textbf{Sample-time reward composition.}
Each transition stores $(\widetilde o_t,\delta_t,\bar r_t,c^{\mathrm{ov}}_t,c^{\mathrm{def}}_t,c^{\mathrm{aux}}_t,\widetilde o_{t+1},\iota_t)$, with the learner observation $\widetilde o_t$ and residual $\delta_t$ of M2, and the terminal indicator $\iota_t$. Its reward is composed only when it is sampled,
\begin{equation}
    \label{eq:compose}
    \widetilde r_t(\lambda)=\bar r_t
    -\lambda_{\mathrm{ov}}T\Delta t\,c^{\mathrm{ov}}_t
    -\lambda_{\mathrm{def}}c^{\mathrm{def}}_t
    {}-\zeta\kappa\, c^{\mathrm{aux}}_t.
\end{equation}
The policy does not observe $\lambda$. A dual update reprices every stored transition. Feature scaling is fixed, and rewards are not normalized. Repricing is approximate because the bootstrap target~\eqref{eq:sac-target} was fitted under earlier multipliers. However, a dual period spans far more gradient steps than the target time constant $1/\tau$ (Table~\ref{tab:setup}), so $\lambda$ is quasi-static for the critics.

\textbf{Prioritized experience replay (PER).}
Transitions are sampled with probability $p_t$ proportional to $(|\Delta_t^{\mathrm{TD}}|+\varepsilon_{\mathrm{PER}})^{\omega}$, where $\Delta_t^{\mathrm{TD}}$ is the temporal-difference error~\cite{schaul2016}. Importance weights $\varpi_t=(|\mathcal B|p_t)^{-\beta}/\max_{t'}(|\mathcal B|p_{t'})^{-\beta}$, with the exponent $\beta$ annealed to full correction (Table~\ref{tab:setup}), scale the losses.

\subsubsection{\textbf{M1: Self-Tuning Constraint Weights}}
\label{ssec:m1}

With $\mathcal{K}_n$ the training days completed in dual window $n$ and $\mathcal{K}'_n\subseteq\mathcal{K}_n$ those with a departure, the violations are
\begin{equation}
\label{eq:pi-v}
\begin{aligned}
v^{\mathrm{ov}}(n)&=\frac{1}{|\mathcal{K}_n|}\sum_{e\in\mathcal{K}_n}\tilde\epsilon^{\mathrm{ov}}_e-\tau^{\mathrm{ov}},\\
v^{\mathrm{def}}(n)&=\epsilon^{\mathrm{usr}}_{\mathrm{train}}-\frac{1}{|\mathcal{K}'_n|}\sum_{e\in\mathcal{K}'_n}\epsilon^{\mathrm{usr}}_e,
\end{aligned}
\end{equation}
namely the mean daily overload above the tightened limit (kWh/day) and the satisfaction shortfall from the training floor. Each multiplier follows a PI controller~\cite{stooke2020},
\begin{subequations}
\label{eq:pi}
\begin{align}
\Xi_q(n) &= \mathrm{clip}_{[0,\,\lambda_{\max}]}
\bigl( \Xi_q(n-1) + K_i\, v^q(n) \bigr),
\label{eq:pi-int}\\
\lambda_q(n) &= \mathrm{clip}_{[0,\,\lambda_{\max}]}
\bigl( K_p\, v^q(n) + \Xi_q(n) \bigr),
\label{eq:pi-out}
\end{align}
\end{subequations}
where $\Xi_q$ integrates past violations with gain $K_i$, the term $K_p v^q$ reacts to the latest one, and $\Xi_q(0) = \lambda_q(0)$ (values in Table~\ref{tab:setup}).

\textbf{Attainability.} With $\tau^{\mathrm{ov}}=0$, the violation $v^{\mathrm{ov}}(n)$ is nonnegative, so the integrator $\Xi_{\mathrm{ov}}$ cannot decrease. It stops growing only when the measured overload vanishes, which requires the unavoidable overload of Remark~\ref{rem:slater}(ii) to be zero. Otherwise, the integral grows toward the cap, where $\lambda_{\mathrm{ov}}$ becomes a fixed price. The residual radius $\rho$~\eqref{eq:projection} matters here. For $\rho<1$, even $\delta_t=-1$ everywhere leaves the command $\max\{a^0_t-\rho,0\}$, so a limit-blind incumbent can force overload that no residual removes. With $\rho\ge1$, the target is attainable (Remark~\ref{rem:slater}(ii)), and the cap is only a safeguard. Sections~\ref{ssec:adaptation} and~\ref{ssec:ablation} report both cases.

\subsubsection{\textbf{M2: Incumbent Warm Start and Residual Anchoring}}
\label{ssec:m2}

The incumbent is the site's existing rule-based dispatcher, such as Round-Robin or CAFAP (Section~\ref{ssec:related}). Its action $a^0_t$ is appended to the observation, $\widetilde o_t = (o_t, a^0_t)$, and the policy outputs a residual~\cite{silver2018residual,johannink2019residual} $\delta_t \in [-1,1]^N$, applied as
\begin{equation}
\label{eq:projection}
a_t \;=\; \operatorname{clip}\!\left(a^0_t + \rho\,\delta_t,\; 0,\; 1\right)
\end{equation}
with residual radius $\rho$ (Table~\ref{tab:setup}). For $\rho\ge1$, every action in $[0,1]$ is reachable on ports that can charge, so the anchor does not bound the action. Instead, it places the origin at the incumbent's command. The residual $\delta_t=0$ reproduces this command, and the initial deterministic policy (last actor layer initialized near zero) nearly reproduces it. Before online learning, the incumbent fills the buffer with $D$ demonstrations~\cite{hester2018,vecerik2017} ($\delta=0$), which sample-time composition keeps correctly priced. Section~\ref{ssec:ablation} evaluates M2.

The incumbent is established, not safe, because a rule such as Round-Robin ignores the transformer limits. M2 is thus an initialization, not safe exploration. No safety mechanism acts in training, where the limit enters only through $\lambda_{\mathrm{ov}}$. Exploration starts wide, so the learner overloads transformers early in training (Section~\ref{ssec:adaptation}).

\subsubsection{\textbf{M3: Grid-Aware Observation and Structured Encoder}}
\label{ssec:m3}

The observation stacks site, transformer and port blocks,
\begin{equation}
\label{eq:obs}
o_t=\bigl[\,o^{\mathrm{g}}_t,\ \{o^{\mathrm{w}}_{w,t}\}_{w\in\mathcal{W}},\ \{o^{\mathrm{c}}_{j,t}\}_{j\in\mathcal{J}}\,\bigr]\in\R^{n_o},
\end{equation}
with $n_o=4+4|\mathcal{W}|+3N$, where
\begin{align}
o^{\mathrm{g}}_t&=\Bigl[\frac{t}{T},\ \frac{P^{\mathrm{set}}_t}{P_{\mathrm{ref}}},\ \frac{P^{\mathrm{tot}}_{t-1}}{P_{\mathrm{ref}}},\ \frac{P^{\mathrm{pot}}_t}{P_{\mathrm{ref}}}\Bigr],
\label{eq:obs-g}\\
o^{\mathrm{w}}_{w,t}&=\frac{\bigl[\overline P^{\mathrm{tr}}_{w,t},\ \overline P^{\mathrm{tr},h}_{w,t},\ P^{\mathrm{nEV}}_{w,t},\ \overline P^{\mathrm{tr}}_{w,t}-P^{\mathrm{nEV}}_{w,t}\bigr]}{\overline P^{\mathrm{nom}}_{w}},
\label{eq:obs-w}\\
o^{\mathrm{c}}_{j,t}&=\bigl[\,d_{j,t},\ \Delta E_{j,t}/E_{\mathrm{sc}},\ \Delta\tau_{j,t}/T\,\bigr].
\label{eq:obs-c}
\end{align}
Here $P^{\mathrm{tot}}_{-1}=0$. Each transformer block is scaled by its limit outside DR events, $\overline P^{\mathrm{nom}}_w=\max_{t}\overline P^{\mathrm{tr}}_{w,t}$. Its second entry, $\overline P^{\mathrm{tr},h}_{w,t}$, is the minimum limit over $\mathcal{H}_t=\{t,\dots,t+h-1\}$. The window length $h\Delta t$ equals the DR notice (Table~\ref{tab:setup}). Because $\mathcal{H}_t$ includes $t$, an event enters the transformer block $h-1$ intervals before it starts. The loading of transformer $w$ is
\begin{equation}
\label{eq:trload}
P^{\mathrm{tr}}_{w,t}=\underbrace{\sum_{j\in\mathcal{J}_w}P^{\mathrm{ch}}_{j,t}}_{P^{\mathrm{EV}}_{w,t}}+\underbrace{P^{L}_{w,t}-P^{PV}_{w,t}}_{P^{\mathrm{nEV}}_{w,t}},
\end{equation}
with $P^{\mathrm{ch}}_{j,t}$ booked battery-side (Section~\ref{ssec:pst}) and $P^{PV}_{w,t}\ge0$, so $P^{\mathrm{nEV}}_{w,t}$ can be negative. The simulator supplies $P^{\mathrm{nEV}}_{w,t}$ of the dispatched interval, a perfect nowcast. Section~\ref{ssec:robust} instead tests the previous interval's value.

With the tightened limit $\widetilde P^{\mathrm{tr}}_{w,t}$ of~\eqref{eq:exact-costs} and the loading~\eqref{eq:trload},
\begin{equation}
\label{eq:4-covbound}
\begin{aligned}
c^{\mathrm{ov}}_t&=\frac{\Delta t}{E_{\mathrm{ref}}}\sum_{w\in\mathcal{W}}\pos{P^{\mathrm{EV}}_{w,t}-H^{m}_{w,t}}\\
&\le\frac{\Delta t}{E_{\mathrm{ref}}}\sum_{w\in\mathcal{W}}\pos{U_{w}(o_t,a_t)-H^{m}_{w,t}},\\
H^{m}_{w,t}&=\widetilde P^{\mathrm{tr}}_{w,t}-P^{\mathrm{nEV}}_{w,t}=\pos{(1-m)\overline P^{\mathrm{tr}}_{w,t}-P^{\mathrm{nEV}}_{w,t}},\\
U_{w}(o_t,a_t)&=\textstyle\sum_{i\in\mathcal{C}_w}\overline P^{\mathrm{cs}}_i\min\bigl\{1,\sum_{j\in\mathcal{J}_i}a_{j,t}u_{j,t}\bigr\},
\end{aligned}
\end{equation}
where $\mathcal{C}_w$ are the stations fed by $w$, and $u_{j,t}=\mathbb{I}\{d_{j,t}>0\}$ is observed. The headroom $H^{m}_{w,t}$ below the tightened limit follows from $o^{\mathrm{w}}_{w,t}$ and $\overline P^{\mathrm{nom}}_w$. By~\eqref{eq:current}--\eqref{eq:chpower}, $U_w$ bounds the power that the commanded currents can deliver behind $w$. Hence $c^{\mathrm{ov}}_t$ is not a function of $o_t$ alone, but it is bounded by a known function of $(o_t,a_t)$. This bound is zero when the commanded load fits the observed headroom. Only the power that the vehicles accept is unobserved. Without~\eqref{eq:obs-w}, $H^{m}_{w,t}$ would depend on the hidden $\xi_t$.

For EV $k$ at port $j$, the features
\[
\Delta E_{j,t}=\sum_{t'=t^{\mathrm{arr}}_k}^{t-1}P^{\mathrm{ch}}_{j,t'}\,\Delta t,\qquad
\Delta\tau_{j,t}=t-t^{\mathrm{arr}}_k,
\]
(zero on an empty port) come from the charger's session record (metered energy and plug-in time), so no $E^{\mathrm{arr}}_k$ is needed. They are scaled by $E_{\mathrm{sc}}$ (Table~\ref{tab:setup}) and $T$. On a real charger, the metered energy is grid-side and thus up to a factor $1/\eta_k$ above $\Delta E_{j,t}$.

The port status
\begin{equation}
\label{eq:4-status}
d_{j,t}=\begin{cases}
0, & \text{no vehicle connected},\\
0.5, & \text{vehicle connected and accepting power},\\
1, & \text{vehicle connected, accepting no power},
\end{cases}
\end{equation}
is the OCPP~1.6 connector state~\cite{ocpp16} (\emph{Available} for $0$, \emph{Charging} or \emph{SuspendedEVSE} for $0.5$, and \emph{SuspendedEV} for $1$). It derives from the IEC~61851-1 control pilot~\cite{iec61851} and is not a state-of-charge reading. In the simulator, $d_{j,t}=1$ means a full battery. On a real site, $d_{j,t}=1$ also covers a self-pausing vehicle and appears only while current is offered. Empty and full ports both draw no current, but only a full one has an open session ($\Delta E_{j,t}>0$). Section~\ref{ssec:robust} flips the status and hides or delays the port data seen by the policy.

Following Deep Sets~\cite{zaheer2017} (shared maps, symmetric pooling), M3 applies three multilayer perceptrons (MLPs), each with two hidden layers of width $n_{\mathrm h}$, to every port ($\chi_{\mathrm c}$), every transformer ($\chi_{\mathrm w}$) and the site block ($\chi_{\mathrm g}$),
\begin{equation}
\label{eq:m3-port}
\begin{aligned}
e_{j,t}&=\chi_{\mathrm c}\bigl([o^{\mathrm c}_{j,t},a^0_{j,t}]\bigr),\quad
\bar e_{w,t}=\frac{\sum_{j\in\mathcal J_w}u_{j,t}\,e_{j,t}}{\max\{1,N^{\mathrm{occ}}_{w,t}\}},\\
q_{w,t}&=\chi_{\mathrm w}\Bigl(\Bigl[o^{\mathrm w}_{w,t},\bar e_{w,t},\frac{N^{\mathrm{occ}}_{w,t}}{N_w}\Bigr]\Bigr),\quad
g_t=\chi_{\mathrm g}\bigl(o^{\mathrm g}_t\bigr).
\end{aligned}
\end{equation}
Here $N^{\mathrm{occ}}_{w,t}=\sum_{j\in\mathcal J_w}u_{j,t}$ counts the occupied ports behind $w$, and $\bar e_{w,t}$ averages their embeddings. The vector $q_{w,t}$ embeds $w$ with its ports, and $g_t$ embeds the site block. The occupied share ($N_w=|\mathcal J_w|$) restores the count that the mean discards, so $q_{w,t}$ can represent the EV load behind $w$ next to its headroom. This share and $a^0_{j,t}$ in $\chi_{\mathrm c}$ are the two load-aware inputs, ablated in Section~\ref{ssec:ablation}. A shared actor head $\chi_{\mathrm a}$ decodes each residual, with $w(j)$ the transformer of port $j$,
\begin{equation}
\label{eq:m3-actor}
\begin{aligned}
(\mu_{j,t},\log\sigma_{j,t})&=\chi_{\mathrm a}\bigl([e_{j,t},q_{w(j),t},g_t,a^0_{j,t}]\bigr),\\
\delta_{j,t}&=c_{j,t}\tanh(\mu_{j,t}+\sigma_{j,t}\nu_{j,t}),
\end{aligned}
\end{equation}
with $\nu_{j,t}\sim\mathcal N(0,1)$ and $c_{j,t}=\mathbb{I}\{d_{j,t}=0.5\}$.
Each twin critic $\ell\in\{1,2\}$ has its own maps of these kinds and a head $\chi^{Q,\ell}_{\mathrm f}$. It pools the port state--residual embeddings per transformer and then over the site,
\begin{equation}
\label{eq:m3-critic}
\begin{aligned}
z^{(\ell)}_{w,t}&=\frac{\sum_{j\in\mathcal J_w}u_{j,t}\,
\chi^{Q,\ell}_{\mathrm c}([o^{\mathrm c}_{j,t},a^0_{j,t},\delta_{j,t}])}
{\max\{1,N^{\mathrm{occ}}_{w,t}\}},\\
b^{(\ell)}_{w,t}&=\chi^{Q,\ell}_{\mathrm w}\Bigl(\Bigl[o^{\mathrm w}_{w,t},z^{(\ell)}_{w,t},\frac{N^{\mathrm{occ}}_{w,t}}{N_w}\Bigr]\Bigr),\\
Q_{\theta_\ell}(\widetilde o_t,\delta_t)&=\chi^{Q,\ell}_{\mathrm f}\Bigl(\Bigl[\chi^{Q,\ell}_{\mathrm g}(o^{\mathrm g}_t),\ \frac{1}{|\mathcal W|}\sum_{w\in\mathcal W}b^{(\ell)}_{w,t}\Bigr]\Bigr).
\end{aligned}
\end{equation}

Every pool skips empty ports. Full ports stay as context with $c_{j,t}=0$. Only ports in $\mathcal J_t^{\mathrm{ctrl}}=\{j:d_{j,t}=0.5\}$ get a residual, a log-probability and an entropy term. Under any consistent relabeling $\varsigma$ of the ports, their incidence and $a^0_t$, the actor satisfies $\delta_\psi(\varsigma\widetilde o)=\varsigma\delta_\psi(\widetilde o)$, the critics are invariant, and~\eqref{eq:projection} is equivariant. The parameter count does not depend on $N$ or $|\mathcal W|$. The actor has $7n_{\mathrm h}^2+23n_{\mathrm h}+2$ parameters, and each critic has $6n_{\mathrm h}^2+22n_{\mathrm h}+1$.

\subsection{TC-CPL Algorithm}
\label{ssec:algorithm}

TC-CPL optimizes the composed reward~\eqref{eq:compose} with SAC~\cite{haarnoja2018}, a maximum-entropy actor-critic method. It keeps a stochastic actor $\pi_\psi$ over the residual $\delta_t$, twin critics $Q_{\theta_\ell}$ and slowly updated targets $Q_{\bar\theta_\ell}$, $\ell\in\{1,2\}$. For a sampled transition, both critics regress toward
\begin{equation}
\label{eq:sac-target}
\begin{aligned}
y_t={}&\widetilde r_t(\lambda)+(1-\iota_t)\gamma\Bigl( \\
&\min_{\ell \in \{1,2\}}Q_{\bar\theta_\ell}(\widetilde o_{t+1},\delta')
-\alpha\log\pi_\psi(\delta'\mid\widetilde o_{t+1})\Bigr),
\end{aligned}
\end{equation}
where $\alpha > 0$ is the entropy temperature, $\delta'\sim\pi_\psi(\cdot \mid \widetilde o_{t+1})$ and $\gamma$ is the discount factor. The critics minimize the importance-weighted squared residual
\begin{equation}
\label{eq:sac-critic}
J_Q(\theta_\ell) = \E_{(\widetilde o_t, \delta_t) \sim \mathcal{B}}
\Bigl[ \tfrac{\varpi_t}{2} \bigl( Q_{\theta_\ell}(\widetilde o_t, \delta_t) - y_t \bigr)^2
\Bigr],
\end{equation}
and the actor minimizes
\begin{equation}
\label{eq:sac-actor}
\begin{aligned}
J_\pi(\psi)=\E_{\widetilde o_t \sim \mathcal{B},\,\delta \sim \pi_\psi}
\Bigl[\varpi_t\bigl(&\alpha\log\pi_\psi(\delta\mid\widetilde o_t)\\
&-\min_{\ell \in \{1,2\}}Q_{\theta_\ell}(\widetilde o_t,\delta)\bigr)\Bigr],
\end{aligned}
\end{equation}
with the reparameterization trick. The minimum over the two critics counters value overestimation, and $\alpha$ is adjusted automatically~\cite{haarnoja2018auto} to hold the policy entropy near $-|\mathcal J_t^{\mathrm{ctrl}}|$. Algorithm~\ref{alg:TC-CPL} summarizes training, and Table~\ref{tab:setup} lists its constants.

\begin{algorithm}[!t]
\caption{TC-CPL training}
\label{alg:TC-CPL}
\begin{algorithmic}[1]
\STATE \textbf{Input:} constants of Table~\ref{tab:setup}
\STATE Initialize $\pi_\psi$, $Q_{\theta_\ell}$, $\bar\theta_\ell \leftarrow \theta_\ell$, $\alpha$, $\Xi_q \leftarrow \lambda_q(0)$; empty buffer $\mathcal{B}$
\STATE Store $D$ incumbent transitions ($\delta=0$) in $\mathcal{B}$
\FOR{each environment step, up to $T_{\mathrm{tot}}$}
  \STATE Draw $\delta_t \sim \pi_\psi(\cdot\,|\,\widetilde o_t)$, apply $a_t$ by \eqref{eq:projection}, store the transition in $\mathcal{B}$
  \STATE Sample a prioritized minibatch; compose $\widetilde r_t(\lambda)$ by \eqref{eq:compose}
  \STATE Step on \eqref{eq:sac-critic}, \eqref{eq:sac-actor} and $\alpha$; update $\bar\theta_\ell$ with rate $\tau$ and the priorities
  \IF{$F$ steps have passed since the last dual update}
    \STATE Update $\lambda$ by \eqref{eq:pi} from the violations \eqref{eq:pi-v}
  \ENDIF
\ENDFOR
\STATE \textbf{Output:} deterministic policy $\delta_{j,t}=c_{j,t}\tanh\mu_{j,t}$
\end{algorithmic}
\end{algorithm}

\textbf{Computational complexity.} Per decision, $\chi_{\mathrm c}$ and $\chi_{\mathrm a}$ run once per port, $\chi_{\mathrm w}$ once per transformer and $\chi_{\mathrm g}$ once. The actor thus needs $O(Nn_{\mathrm h}^2+|\mathcal W|n_{\mathrm h}^2+|\mathcal W|Nn_{\mathrm h})$ operations and $O(Nn_{\mathrm h}+|\mathcal W|n_{\mathrm h})$ activations per sample. The critics scale alike, and a training update multiplies these costs by the minibatch size $B$. The last term of the operation count is the masked-mean pooling, a dense $|\mathcal W|\times N$ product. A segment sum would reduce it to $O(Nn_{\mathrm h})$. The observation, incumbent and projection cost $O(N+|\mathcal W|)$. Section~\ref{ssec:compute} reports measured costs.

\section{Validation}
\label{sec:validation}

We evaluate TC-CPL in two EV2Gym scenarios~\cite{ev2gym} in which the transformer limit binds.

\subsection{Experimental Setup and Evaluation Metrics}
\label{ssec:setup}

\begin{table*}[!t]
\centering
\caption{Experimental settings (a)~per scenario and (b)~shared by both scenarios and TC-CPL variants in all main runs. Ranges span transformers and evaluation days.}
\label{tab:setup}
\scriptsize
\def\tsrc{}%
\def\tshd#1{\begin{tabular}[b]{@{}>{\bfseries\boldmath}c@{}}#1\end{tabular}}%
\setlength{\tabcolsep}{0pt}%
\begin{tabular*}{\textwidth}{@{\extracolsep{\fill}}>{\tsrc}l*{13}{>{\tsrc}c}@{}}
\toprule
\textbf{(a) Per scenario} & & \multicolumn{4}{c}{\tsrc\bfseries Transformers} & \multicolumn{3}{c}{\tsrc\bfseries Daily demand} & \multicolumn{2}{c}{\tsrc\bfseries Day} & \multicolumn{3}{c}{\tsrc\bfseries Learner}\\
\cmidrule(lr){3-6}\cmidrule(lr){7-9}\cmidrule(lr){10-11}\cmidrule(l){12-14}
\textbf{Scenario} & \tshd{Ports $N$\\(stations)} & \tshd{$|\mathcal W|$\\(with ports)} & \tshd{Ports\\each} & \tshd{Limit\\outside DR,\\$\overline P^{\mathrm{nom}}_w$ (kW)} & \tshd{Lowest\\in DR\\(kW)\textsuperscript{b}} & \tshd{Non-EV\\peak /\\limit} & \tshd{EV\\sessions\\per day} & \tshd{Setpoint\\/ EV\\energy} & \textbf{Start} & \textbf{Days} & \tshd{$P_{\mathrm{ref}}$\\(kW)} & \tshd{$E_{\mathrm{ref}}$\\(kWh)} & \tshd{$n_o$} \\
\midrule
\texttt{PublicPST-S} & $20$ ($20$) & $1$ ($1$) & $20$ & $42$ & $20.4$--$38.0$ & $0.29$--$0.79$ & $6$--$21$ & $1.8$ & 05:00 & weekdays & $\phantom{3{,}}221.7$ & $\phantom{7}5{,}321$ & $\phantom{2{,}2}68$\\
\texttt{cityNL} & $600$ ($300$) & $122$ ($93$) & $2$--$28$ & $95$--$760$\textsuperscript{a} & $56.6$--$656.7$ & $0.20$--$0.78$ & $981$--$1{,}112$ & $1.1$ & 00:00 & all days & $3{,}325.5$ & $79{,}813$ & $2{,}292$\\
\midrule
\end{tabular*}

\begin{tabular*}{\textwidth}{@{\extracolsep{\fill}}ll@{}}
\begin{tabular}[t]{@{}>{\tsrc}l@{\quad}>{\tsrc}l@{\quad}>{\tsrc}l@{}}
\textbf{(b) Shared} & \multicolumn{2}{c@{}}{\tsrc\bfseries Constraints}\\
\cmidrule(l){2-3}
& \textbf{Training} & \textbf{Evaluation}\\
\cmidrule(l){2-2}\cmidrule(l){3-3}
Transformer limit & $\widetilde P^{\mathrm{tr}}_{w,t}$, back-off $m=0.05$ & $\overline P^{\mathrm{tr}}_{w,t}$, raw, no band\\
Overload threshold & $\tau^{\mathrm{ov}}=0$ ($d^{\mathrm{ov}}=0$) & $\epsilon^{\mathrm{ov}}_{\max}=0$~kWh/day\\
Satisfaction floor & $\epsilon^{\mathrm{usr}}_{\mathrm{train}}=0.98$ ($d^{\mathrm{def}}=0.02$) & $\epsilon^{\mathrm{usr}}_{\min}=0.95$ ($\epsilon^{\mathrm{def}}_{\max}=0.05$)\\
\end{tabular}
&
\begin{tabular}[t]{@{}>{\tsrc}l@{\enspace}>{\tsrc}l@{}}
\multicolumn{2}{@{}c@{}}{\tsrc\bfseries Charger, EV and day}\\
\cmidrule{1-2}
$\overline I^{\mathrm{cs}}_i$, $\overline P^{\mathrm{cs}}_i$ & $16$~A, $11.09$~kW ($V_j=400$~V, $\phi_j=3$)\\
$\eta_k$ & $[0.95,1]$\\
Taper & above a state of charge of $0.7$--$0.9$\\
$\mathrm{SoC}^{\star}_k$ & $1$; every EV departs within its day\\
$T\times\Delta t$ & $96\times15$~min ($24$~h)\\
$h\Delta t$ & $60$~min notice; an event enters $o^{\mathrm{w}}_{w,t}$ $45$~min before it starts\\
\end{tabular}
\\
\end{tabular*}

\vspace{6pt}
\begin{tabular*}{\textwidth}{@{\extracolsep{\fill}}lll@{}}
\begin{tabular}[t]{@{}>{\tsrc}l@{\enspace}>{\tsrc}l@{}}
\multicolumn{2}{@{}c@{}}{\tsrc\bfseries M1}\\
\cmidrule{1-2}
$K_p$, $K_i$ & $5$, $2$ (PI gains)\\
$\lambda_{\max}$ & $500$ (cap on $\lambda_q$ and $\Xi_q$)\\
$\lambda(0)=\Xi(0)$ & $(10,5)$\\
$F$ & $10^{4}$ steps (about $104$ days)\\
\multicolumn{2}{@{}l@{}}{\tsrc\itshape Fixed-price TC-CPL}\\
$(\lambda_{\mathrm{ov}},\lambda_{\mathrm{def}})$ & $(100,50)$\\
\end{tabular}
&
\begin{tabular}[t]{@{}>{\tsrc}l@{\enspace}>{\tsrc}l@{}}
\multicolumn{2}{@{}c@{}}{\tsrc\bfseries Soft actor-critic and prioritized replay}\\
\cmidrule{1-2}
$\gamma$, initial $\alpha$ & $1$, $1$\\
Adam step size & $3\times10^{-4}$ (actor, critics, $\alpha$)\\
$B$, $|\mathcal{B}|$ & $256$, $2\times10^{5}$ (replay capacity)\\
$\tau$ (Polyak), $1/\tau$ & $0.005$, $200$\\
Environments & $32$ ($1$ gradient step per transition)\\
$\omega$, $\varepsilon_{\mathrm{PER}}$; $\beta$ & $0.6$, $10^{-3}$; $0.4\to1$, linear in training\\
\end{tabular}
&
\begin{tabular}[t]{@{}>{\tsrc}l@{\enspace}>{\tsrc}l@{}}
\multicolumn{2}{@{}c@{}}{\tsrc\bfseries M2, M3 and training budget}\\
\cmidrule{1-2}
$D$ & $10^{5}$ ($1{,}042$ days, $\delta=0$)\\
$\rho$ & $1$\\
$n_{\mathrm h}$ & $128$ (SiLU; one hidden layer per head)\\
$E_{\mathrm{sc}}$, $h$ & $100$~kWh, $4$\\
$\zeta$, $\kappa$ & $25$, $1\to0$ (linear, by $0.8\,T_{\mathrm{tot}}$)\\
$T_{\mathrm{tot}}$ & $10^{6}$ steps per learner and seed\\
\end{tabular}
\\
\bottomrule
\end{tabular*}

\vspace{2pt}
\parbox{\textwidth}{\raggedright\tsrc\textsuperscript{a}Ratings are $42$~kW (\texttt{PublicPST-S}) and $4\times100$, $44\times250$, $10\times630$ and $64\times800$~kVA at $0.95$~kW/kVA (\texttt{cityNL}, median limit $760$~kW). \textsuperscript{b}The demand-response (DR) event is redrawn daily (\texttt{PublicPST-S}) or the same every day (\texttt{cityNL}). A $10$-kW band is $23.8\%$ of the \texttt{PublicPST-S} limit and $1.3$--$10.5\%$ of the \texttt{cityNL} limits. The non-EV load $P^{\mathrm{nEV}}_{w,t}$ is household load, peaking near $0.55$ of the limit, minus PV. \texttt{cityNL} links public stations to a Dutch $123$-bus feeder (AC power flow) whose substations are rated at least $1/0.55$ times their nodes' peak demand. Sessions come from ElaadNL~\cite{elaad}. Every evaluation day and all $500$ sampled \texttt{PublicPST-S} training days have departures ($3$--$21$ per \texttt{PublicPST-S} day), so no empty day dilutes the window averages of M1. As $\eta_k\ge0.95=1-m$, grid-side values, a charger's metered energy included, exceed the reported battery-side ones by at most $5.3\%$.}
\end{table*}

\subsubsection{Scenarios and controllers}
Training uses the tightened limit and floor, and evaluation uses the raw limit and the operating requirements (Table~\ref{tab:setup}(b)). All reported values, tracking included, are battery-side (Section~\ref{ssec:pst}). \emph{Self-tuning TC-CPL}, or simply TC-CPL, is the proposed method, and \emph{Fixed-price TC-CPL} is its M1 ablation (Table~\ref{tab:methods}). Its hand-set prices were never tuned. Both variants share the shaping term~\eqref{eq:3-hinge} and the Round-Robin incumbent, and Section~\ref{ssec:ablation} also tests CAFAP. No baseline was tuned. Other methods of Table~\ref{tab:positioning} target different settings or information and were not rerun. CPO, Fixed-price TC-CPL and the $K_p=K_i$ variant stand in for trust-region, fixed-penalty and augmented-Lagrangian methods. The offline optimum solves~\eqref{eq:offline} for each evaluation day (checks in the note to Table~\ref{tab:methods}).

\begin{table*}[!t]
\centering
\caption{Compared controllers and evaluation protocol, the same in both scenarios.}
\label{tab:methods}
\scriptsize
\setlength{\tabcolsep}{2pt}
\begin{tabular}{@{}l>{\raggedright\arraybackslash}p{2.9cm}>{\raggedright\arraybackslash}p{3.6cm}>{\raggedright\arraybackslash}p{4.3cm}>{\raggedright\arraybackslash}p{2.3cm}>{\raggedright\arraybackslash}p{2.25cm}@{}}
\toprule
Controller & Information & Action; network & Constraint handling & Warm start; replay & Budget; tuning \\
\midrule
CAFAP & occupancy & $a_{j,t}=1$ on all ports & none & -- & rule \\
Round-Robin & $P^{\mathrm{set}}_t$, port status & $\lceil P^{\mathrm{set}}_t/\bar P_{\mathrm{port}}\rceil$ EVs in turn & none & M2 incumbent & rule \\
DDPG-PST & PST state, normalized & direct; MLP $128$-$128$/$64$-$64$ & none (tracking reward only) & uniform, $10^6$ & $10^6\times5$; as~\cite{yilmaz2024} \\
PPO-PST & PST state, normalized & direct; MLP $128$-$128$ & none (tracking reward only) & on-policy & $1.016\times10^6\times5$ \\
CPO & $o_t$ (no $a^0_t$), normalized & direct; MLP $256$-$128$-$256$ & trust region; raw overload $\le0$ & on-policy & $1.008\times10^6\times5$ \\
Self-tuning TC-CPL & $\widetilde o_t=(o_t,a^0_t)$ & residual, $\rho=1$; M3, load-aware & $c^{\mathrm{ov}}_t$ above $\widetilde P^{\mathrm{tr}}_{w,t}$, $c^{\mathrm{def}}_t$ at $0.98$; PI law (M1) from $(10,5)$ & $10^5$ Round-Robin steps (M2); PER & $10^6\times5$; Table~\ref{tab:setup} \\
Fixed-price TC-CPL & same & same & same, fixed $(100,50)$, \textbf{no M1} & same & $10^6\times5$; hand-set \\
Offline optimum & hindsight of the day & QP~\eqref{eq:offline}, replayed & hard limit per interval; no $\epsilon^{\mathrm{usr}}$ floor & -- & one QP per day \\
\bottomrule
\end{tabular}

\vspace{2pt}
\parbox{\textwidth}{\raggedright
$\bar P_{\mathrm{port}}$ is the port rating, $11.09$ (\texttt{PublicPST-S}) or $5.54$~kW (\texttt{cityNL}).
\textit{Baselines.} DDPG-PST~\cite{lillicrap2016} and PPO-PST~\cite{schulman2017} use Stable-Baselines3~\cite{raffin2021}. CPO, our implementation of~\cite{achiam2017}, has the tracking reward and no service term. All its $123$ updates per \texttt{cityNL} seed ($89$--$116$ of $123$ on \texttt{PublicPST-S}) were recovery steps, so its low service reflects this configuration, not the method. In the normalized reward, Fixed-price $\lambda_{\mathrm{ov}}=100$ charges $0.45$ (\texttt{PublicPST-S}) or $0.030$ (\texttt{cityNL}) per kWh of overload.
\textit{Offline optimum.} CVXPY~\cite{diamond2016cvxpy} with Clarabel~\cite{goulart2024clarabel} solves a \texttt{cityNL} day ($20{,}487$--$24{,}066$ variables) to default tolerances ($10^{-8}$ on duality gap and feasibility) in $0.18$~s on average. Adding either satisfaction floor changes $J^{\star}_0$ by under $2\times10^{-9}$ (relative) where attainable. Replayed, the squared error exceeds $J^{\star}_0$ by $6.6\%$ (\texttt{PublicPST-S}) and $57\%$ (\texttt{cityNL}) on average. In hindsight, a linear program attains zero overload with $\epsilon^{\mathrm{usr}}\ge0.95$ on every evaluation day, and with $\epsilon^{\mathrm{usr}}\ge0.98$ except on three \texttt{PublicPST-S} days, where one vehicle, not the transformer, is the limit. The tightened limit was not tested in hindsight.
\textit{Service.} We report delivered energy $E^{\mathrm{del}}$, the shares of EVs with $s_k<0.95$ and $s_k<0.80$, the 5th percentile of $s_k$, the undelivered share of all requested energy and the unmet demand $100\bigl(1-|\mathcal{E}|^{-1}\sum_{k}E^{\mathrm{dep}}_k/\widehat E_k\bigr)$, with departure energy $E^{\mathrm{dep}}_k$ and its full-power value $\widehat E_k$ (whole stay, capped at capacity).
\textit{Statistics.} Learners (seeds $42$--$46$) run final checkpoints deterministically, and rules and the offline optimum run once. Demonstrations exclude evaluation days, but training is expected to draw about one per run. Paired intervals are two-level percentile bootstraps over $500$ seed-days, approximate and likely narrow at the seed level. A day is overloaded when $\epsilon^{\mathrm{ov}}_{0}>10^{-3}$~kWh ($10^{-9}$ in Table~\ref{tab:ablation}).}
\end{table*}

\subsubsection{Metrics and statistics}
Because $\epsilon^{\mathrm{usr}}$ can hide poorly served vehicles, we also report per-vehicle service metrics (defined in the note to Table~\ref{tab:methods}). Raw overload $\epsilon^{\mathrm{ov}}$~\eqref{eq:overload} is the headline safety metric. The banded overload
\begin{equation}
\label{eq:reported-overload}
\epsilon^{\mathrm{ov}}_{b}
=\Delta t\sum_{t\in\mathcal T}\sum_{w\in\mathcal W}
\pos{P^{\mathrm{tr}}_{w,t}-\overline P^{\mathrm{tr}}_{w,t}-b},
\end{equation}
is only a sensitivity check, since a fixed band differs in meaning across transformers (Table~\ref{tab:setup}). The violation frequency is the share of overloaded seed-days,
\begin{equation}
\label{eq:violation-frequency}
\widehat p_b=\frac{1}{S\,N_{\mathrm{eval}}}
\sum_{s=1}^{S}\sum_{d=1}^{N_{\mathrm{eval}}}
\mathbb I\{\epsilon^{\mathrm{ov}}_{b,s,d}>0\},
\end{equation}
over $S$ seeds and $N_{\mathrm{eval}}$ days. We also count violating transformer-intervals and report the worst single-interval excess $\max_{s,d,w,t}\pos{P^{\mathrm{tr}}_{w,t}-\overline{P}^{\mathrm{tr}}_{w,t}}$, in kW and relative to $\overline P^{\mathrm{nom}}_w$. All controllers and seeds face the same days. A difference is \emph{resolved} when its paired $95\%$ bootstrap interval over seeds and days excludes zero.

\subsection{City-Scale District (\texttt{cityNL})}
\label{ssec:citynl}

\begin{table*}[!t]
\centering
\caption{Evaluation on $100$ shared days, as mean$\pm$standard deviation over five per-seed means. Max excess is the worst interval per seed. Each $\Delta$ is Self-tuning TC-CPL minus the named controller, paired over the same days and seeds (percentage points for \%), with its $95\%$ bootstrap interval below ($^{*}$ if it excludes zero).}
\label{tab:results}
\scriptsize
\newcommand{\sig}{\rlap{$^{*}$}}
\newcommand{\ci}[1]{[#1]}
\setlength{\tabcolsep}{3pt}\setlength{\aboverulesep}{0.3ex}\setlength{\belowrulesep}{0.55ex}
\newcommand{\scen}[1]{\smash{\raisebox{3.5pt}{\rotatebox[origin=c]{90}{\footnotesize\texttt{#1}}}}}
\begin{tabular}{@{}c@{\hspace{3pt}}lrrrrrrrr@{}}
\toprule
 & & Tracking & \multicolumn{4}{c}{Service} & \multicolumn{3}{c}{Overload} \\
\cmidrule(lr){3-3}\cmidrule(lr){4-7}\cmidrule(l){8-10}
 & Controller & $|\epsilon_{\mathrm{tr}}|$ & $\epsilon^{\mathrm{usr}}$ & $E^{\mathrm{del}}$ & Unmet & EVs $s_k{<}0.95$ & $\epsilon^{\mathrm{ov}}$ & Seed-days & Max excess \\
 & & kWh/d & & kWh/d & \% & \% & kWh/d & \% & kW \\
\midrule
 & CAFAP & 5{,}727 & 0.997 & 11{,}919 & 0.14 & 1.9 & 3.9807 & 76.0 & 41.54 \\
 & Round-Robin & 3{,}411 & 0.959 & 9{,}891 & 3.94 & 23.9 & 3.5932 & 77.0 & 34.20 \\
 & DDPG-PST & 7{,}756$\pm$240 & 0.823$\pm$0.012 & 6{,}370$\pm$402 & 17.61$\pm$1.23 & 47.6$\pm$3.1 & 1.3342$\pm$0.3913 & 54.8$\pm$8.6 & 23.92$\pm$4.44 \\
 & PPO-PST & 7{,}502$\pm$73 & 0.845$\pm$0.002 & 5{,}846$\pm$78 & 15.41$\pm$0.22 & 58.5$\pm$0.8 & 0.9819$\pm$0.1133 & 59.0$\pm$1.2 & 17.44$\pm$3.93 \\
 & CPO & 12{,}036$\pm$48 & 0.686$\pm$0.002 & 1{,}269$\pm$48 & 31.27$\pm$0.23 & 93.7$\pm$0.3 & 0.0041$\pm$0.0047 & 3.4$\pm$3.5 & 0.58$\pm$0.50 \\
& Fixed-price TC-CPL & 2{,}526$\pm$197 & 0.988$\pm$0.002 & 11{,}514$\pm$94 & 1.01$\pm$0.22 & 6.4$\pm$1.0 & 0.0019$\pm$0.0042 & 0.6$\pm$0.9 & 0.67$\pm$1.48 \\
\scen{cityNL} & Self-tuning TC-CPL & 2{,}494$\pm$67 & 0.989$\pm$0.003 & 11{,}510$\pm$138 & 0.99$\pm$0.30 & 6.3$\pm$1.5 & 0.0017$\pm$0.0039 & 0.4$\pm$0.9 & 0.64$\pm$1.43 \\
 & Offline optimum & 1{,}562 & 0.994 & 11{,}740 & 0.41 & -- & 0.0000 & 0.0 & -- \\
\cmidrule(l){2-10}
 & $\Delta$ vs.\ Fixed-price & $-$32 & $+$0.0002 & $-$4 & $-$0.02 & $-$0.1 & $-$0.0001 & $-$0.2 & -- \\[-1.5pt]
 & & \ci{$-$217, $+$124} & \ci{$-$0.0027, $+$0.0029} & \ci{$-$138, $+$123} & \ci{$-$0.30, $+$0.28} & \ci{$-$1.4, $+$1.3} & \ci{$-$0.0090, $+$0.0092} & \ci{$-$2.0, $+$1.6} & \\
 & $\Delta$ vs.\ Round-Robin & $-$917\sig & $+$0.0293\sig & $+$1{,}618\sig & $-$2.95\sig & $-$17.7\sig & $-$3.5914\sig & $-$76.6\sig & -- \\[-1.5pt]
 & & \ci{$-$975, $-$855} & \ci{$+$0.0273, $+$0.0318} & \ci{$+$1{,}521, $+$1{,}735} & \ci{$-$3.20, $-$2.74} & \ci{$-$18.9, $-$16.6} & \ci{$-$4.4328, $-$2.8323} & \ci{$-$84.6, $-$68.0} & \\
\midrule
 & CAFAP & 143.0 & 0.998 & 130.9 & 0.00 & 1.2 & 5.2332 & 63.0 & 27.94 \\
 & Round-Robin & 100.0 & 0.974 & 122.7 & 2.46 & 12.5 & 1.8577 & 49.0 & 19.33 \\
 & DDPG-PST & 129.6$\pm$0.8 & 0.995$\pm$0.002 & 129.9$\pm$0.3 & 0.34$\pm$0.16 & 2.5$\pm$0.6 & 5.5810$\pm$0.1764 & 67.6$\pm$2.4 & 32.24$\pm$3.63 \\
 & PPO-PST & 151.0$\pm$2.2 & 0.873$\pm$0.007 & 76.4$\pm$2.9 & 12.58$\pm$0.71 & 48.9$\pm$2.3 & 0.3116$\pm$0.0704 & 18.0$\pm$1.6 & 11.94$\pm$3.01 \\
 & CPO & 210.8$\pm$1.2 & 0.691$\pm$0.007 & 12.4$\pm$1.3 & 30.82$\pm$0.75 & 93.6$\pm$1.8 & 0.0000$\pm$0.0000 & 0.0$\pm$0.0 & 0.00$\pm$0.00 \\
& Fixed-price TC-CPL & 123.5$\pm$6.0 & 0.975$\pm$0.010 & 116.9$\pm$5.2 & 2.37$\pm$1.02 & 15.0$\pm$5.4 & 0.0004$\pm$0.0007 & 0.4$\pm$0.5 & 0.15$\pm$0.26 \\
\scen{PublicPST-S} & Self-tuning TC-CPL & 122.2$\pm$3.7 & 0.985$\pm$0.003 & 122.4$\pm$1.7 & 1.30$\pm$0.35 & 9.0$\pm$1.9 & 0.0014$\pm$0.0020 & 0.6$\pm$0.9 & 0.48$\pm$0.72 \\
 & Offline optimum & 94.2 & 0.994 & 128.5 & 0.38 & -- & 0.0000 & 0.0 & -- \\
\cmidrule(l){2-10}
 & $\Delta$ vs.\ Fixed-price & $-$1.3 & $+$0.0107\sig & $+$5.5\sig & $-$1.08\sig & $-$6.0\sig & $+$0.0011 & $+$0.2 & -- \\[-1.5pt]
 & & \ci{$-$5.4, $+$1.5} & \ci{$+$0.0022, $+$0.0202} & \ci{$+$1.1, $+$10.5} & \ci{$-$2.03, $-$0.22} & \ci{$-$11.3, $-$1.1} & \ci{$-$0.0013, $+$0.0049} & \ci{$-$1.4, $+$1.8} & \\
 & $\Delta$ vs.\ Round-Robin & $+$22.2\sig & $+$0.0113\sig & $-$0.3 & $-$1.17\sig & $-$3.5\sig & $-$1.8563\sig & $-$48.4\sig & -- \\[-1.5pt]
 & & \ci{$+$18.1, $+$26.1} & \ci{$+$0.0052, $+$0.0172} & \ci{$-$3.2, $+$2.2} & \ci{$-$1.76, $-$0.54} & \ci{$-$6.4, $-$0.3} & \ci{$-$2.6518, $-$1.1612} & \ci{$-$58.0, $-$38.8} & \\
\bottomrule
\end{tabular}

\vspace{3pt}
{\setlength{\tabcolsep}{4pt}%
\begin{tabular}{@{}lccccc@{}}
TC-CPL vs.\ Round-Robin & Violating intervals & Overall worst excess, kW (\% of limit) & EVs $s_k{<}0.80$ (\%) & 5th percentile of $s_k$ & Undelivered share (\%) \\
\midrule
\texttt{cityNL} & $2$ vs.\ $200$ & $3.2$ ($1.35\%$) vs.\ $34.2$ & $1.8$ vs.\ $6.2$ & $0.924$ vs.\ $0.780$ & $4.5$ vs.\ $18.0$ \\
\texttt{PublicPST-S} & $3$ vs.\ $154$ & $1.6$ ($3.9\%$) vs.\ $19.3$ ($46\%$) & $2.1$ vs.\ $5.2$ & $0.932$ vs.\ $0.879$ & $5.9$ vs.\ $6.8$ \\
\end{tabular}\par\vspace{2pt}
\parbox{\textwidth}{Intervals are counted in $500$ seed-days (TC-CPL) or $100$ days (Round-Robin). On \texttt{cityNL}, TC-CPL delivers $98.0\%$ of the offline optimum's energy. Grid-side on \texttt{PublicPST-S} (conversion losses about $2.5\%$ of $E^{\mathrm{del}}$), overload rises to $2.126$~kWh/day for Round-Robin and at most from $0.008$ to $0.014$ for $2\times10^5$-step TC-CPL policies.}}
\end{table*}

\begin{figure*}[!t]
\centering
\includegraphics[width=0.9\textwidth]{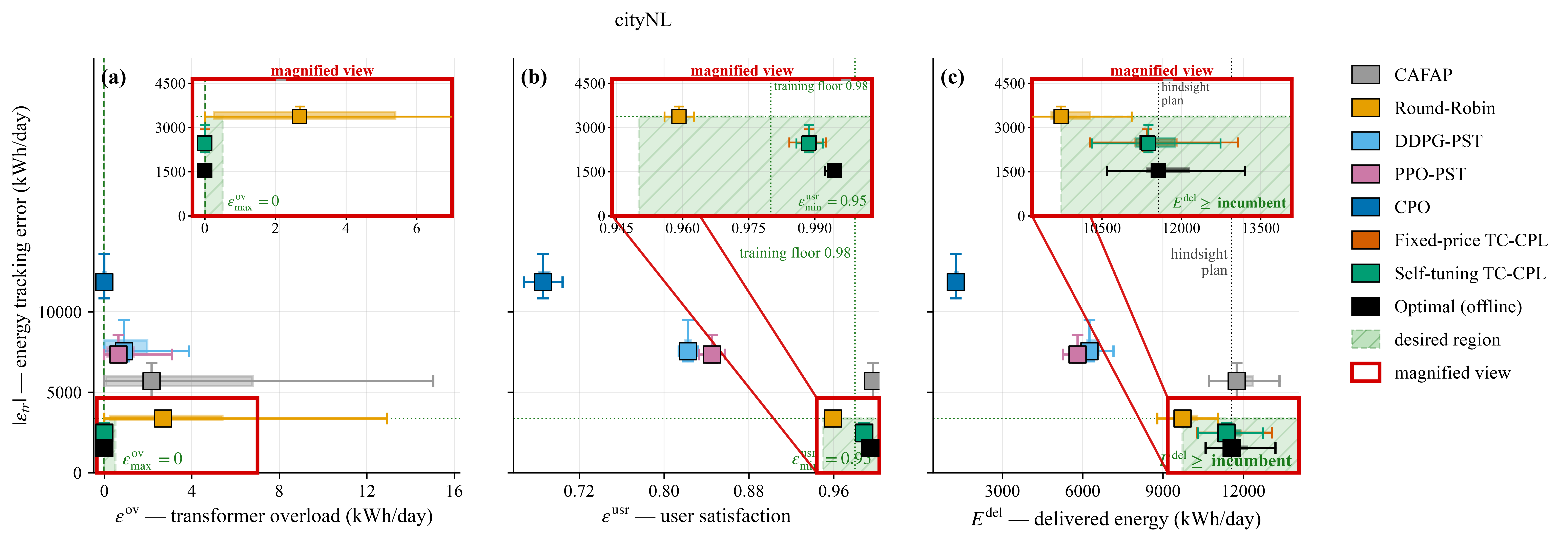}\\[1pt]
\includegraphics[width=0.9\textwidth]{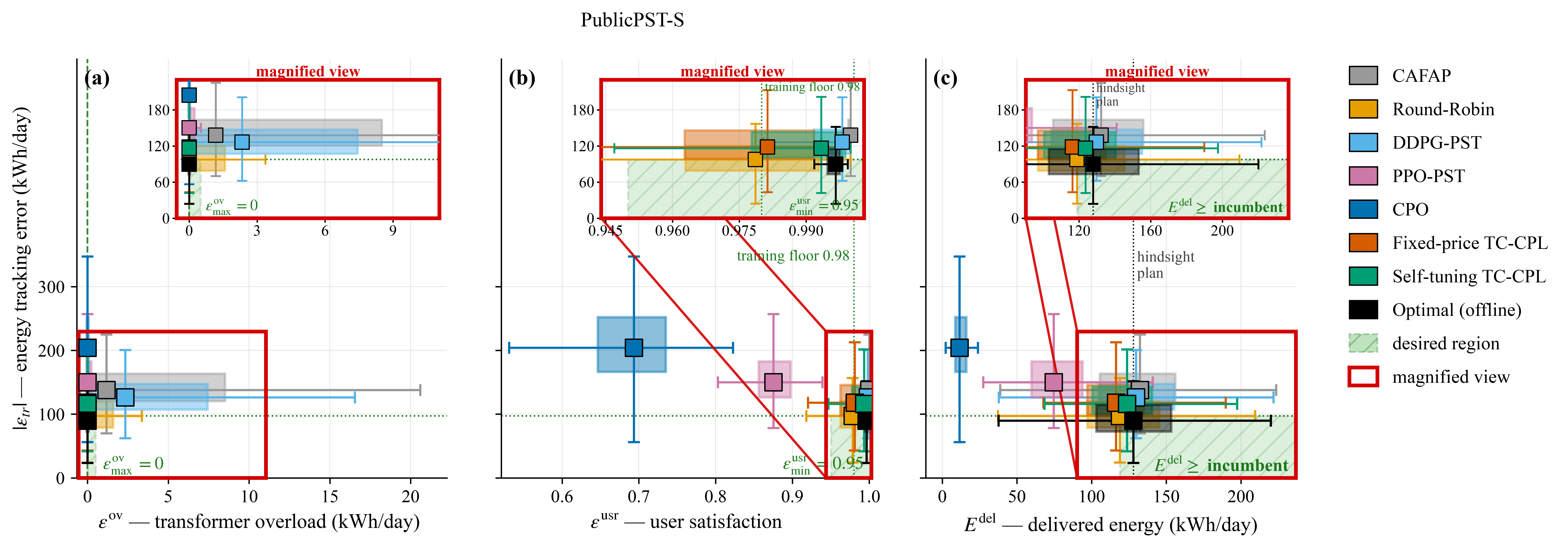}
\caption{Tracking error $|\epsilon_{\mathrm{tr}}|$ against (a)~raw overload, (b)~mean satisfaction and (c)~delivered energy on \texttt{cityNL} (top) and \texttt{PublicPST-S} (bottom). Squares, boxes and whiskers give the median, interquartile range (IQR) and $1.5$~IQR of daily values, seed-averaged for learners. Hatched desired regions pair tracking no worse than Round-Robin's median with (a)~zero overload (drawn $0.5$~kWh/day wide), (b)~$\epsilon^{\mathrm{usr}}\ge0.95$ or (c)~$E^{\mathrm{del}}$ at least Round-Robin's median. On \texttt{cityNL}, only the TC-CPL variants (coinciding) and ``Optimal (offline)'' have medians in all three regions. On \texttt{PublicPST-S}, Round-Robin lies on the boundary by construction (zero median overload, yet overload on $49\%$ of days), and Self-tuning TC-CPL has near-zero overload and the highest satisfaction among limit-respecting online controllers but a median tracking error $19\%$ above Round-Robin's.}
\label{fig:design_city}
\label{fig:design_public}
\end{figure*}

Against Round-Robin (Table~\ref{tab:results}), Self-tuning TC-CPL cuts the raw overload by more than three orders of magnitude and the overall worst excess by one, so overloaded seed-days and violating intervals become rare. It also tracks about a quarter better, delivers about a sixth more energy and raises the mean satisfaction. All three gains are resolved and hold in every seed. Every per-vehicle service metric also improves, and the undelivered share falls to a quarter of Round-Robin's (note to Table~\ref{tab:results}). The two TC-CPL variants are statistically indistinguishable (no interval excludes zero), and their glyphs coincide (Fig.~\ref{fig:design_city}, top).

\begin{figure*}[!t]
\centering
\includegraphics[width=0.9\textwidth]{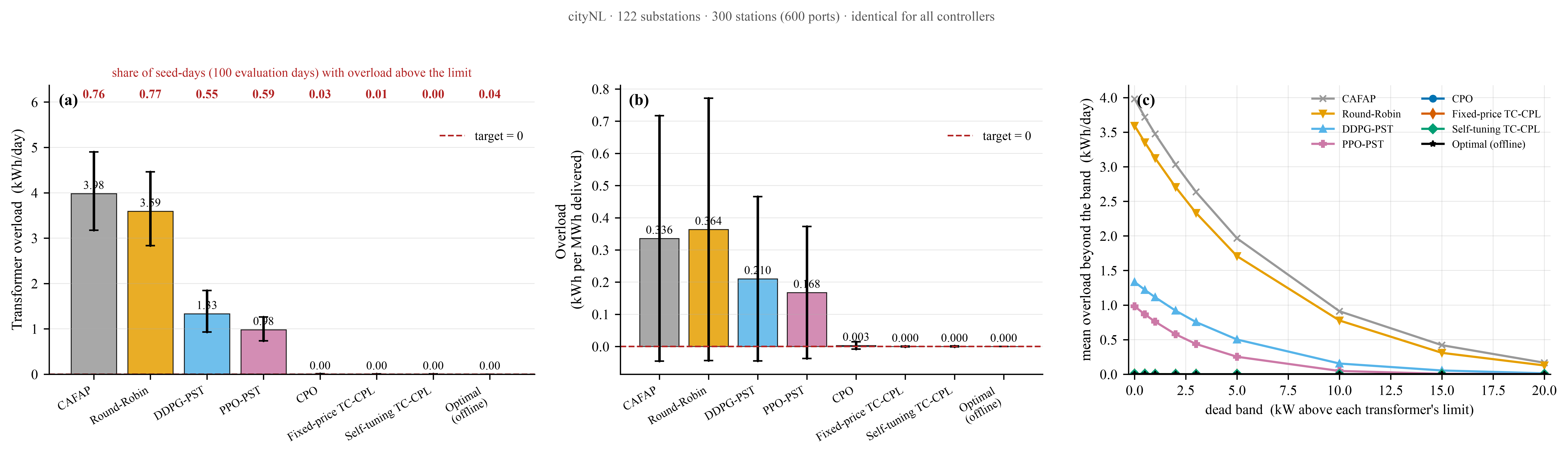}
\caption{Overload on \texttt{cityNL}. (a)~Mean daily raw overload ($95\%$ bootstrap intervals) and share of overloaded seed-days (red). (b)~Overload per MWh delivered, mean $\pm1$ standard deviation over days (hence bars below zero). (c)~Overload beyond a band of $b$~kW, at most $0.004$~kWh/day for both TC-CPL variants at every $b$. The $0.04$ share of ``Optimal (offline)'' is solver residue.}
\label{fig:safety}
\end{figure*}

\begin{figure*}[!t]
\centering
\includegraphics[width=0.9\textwidth]{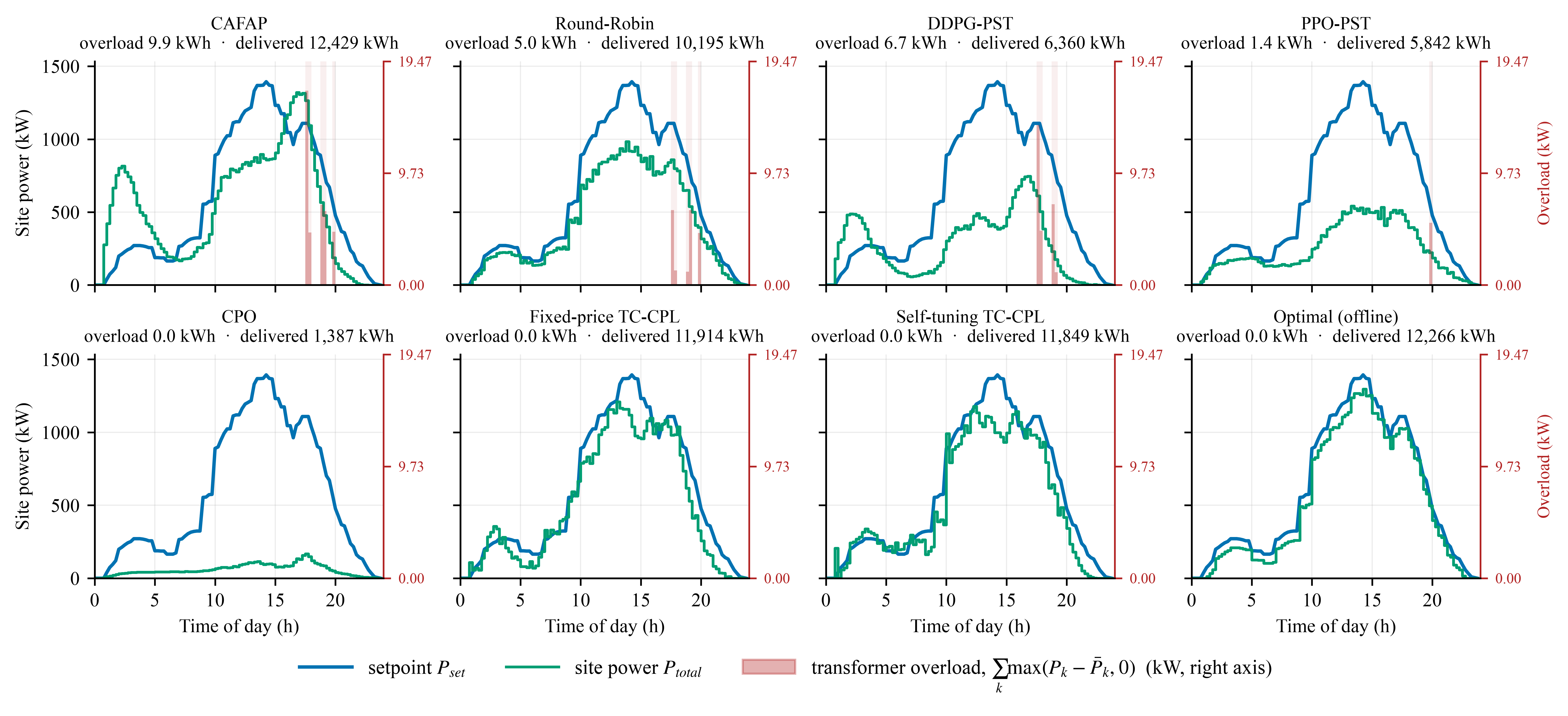}
\caption{Setpoint $P^{\mathrm{set}}_t$ (blue), site power $P^{\mathrm{tot}}_t$ (green) and excess over the DR-aware limits summed over substations (red) on a \texttt{cityNL} day outside the evaluation set (seed $27$; seed-$42$ models). Panel titles give the daily overload and delivered energy. Self-tuning TC-CPL tracks without overload and delivers $96.6\%$ of the offline optimum's energy, $65$~kWh less than Fixed-price TC-CPL.}
\label{fig:setpoint}
\end{figure*}

Every other online controller misses a requirement by far. CAFAP ignores the setpoint and overloads on most days. DDPG-PST and PPO-PST miss the satisfaction requirement, track more than twice as badly as Round-Robin and still overload. CPO nearly avoids overload only by barely charging, and per MWh delivered it overloads about twenty times as much as TC-CPL (Fig.~\ref{fig:safety}(b)). The overload ranking holds at every absolute band (Fig.~\ref{fig:safety}(c)). TC-CPL delivers nearly the offline optimum's energy but tracks markedly worse. Fig.~\ref{fig:setpoint} shows one day.

\subsection{Stressed Single-Transformer Facility (\texttt{PublicPST-S})}
\label{ssec:publicpst}

On \texttt{PublicPST-S}, the limit binds. Round-Robin overloads on about half of the days, by up to nearly half the limit, and CAFAP and DDPG-PST overload even more often.

Self-tuning TC-CPL keeps the overload near zero, with one violating interval on each of its few overloaded seed-days and a small worst excess (note to Table~\ref{tab:results}). At similar delivered energy, it serves better than Round-Robin on every other service metric but tracks about a fifth worse, and Round-Robin keeps most of this lead when projected to zero overload (Section~\ref{ssec:robust}). No learner reaches the desired region (Fig.~\ref{fig:design_public}, bottom), and the tracking gap to the offline optimum is TC-CPL's main shortfall here.

Here, Self-tuning TC-CPL also serves better than Fixed-price TC-CPL, with resolved gains in all four service columns of Table~\ref{tab:results} and unresolved overload and tracking differences. Fixed-price TC-CPL serves no better than Round-Robin and delivers less energy. Recounted grid-side, with conversion losses of a few percent of the delivered energy, Round-Robin's overload rises by about a seventh, and that of shorter-trained TC-CPL policies stays near zero (note to Table~\ref{tab:results}).

\subsection{Constraint Adaptation and Learning Curves}
\label{ssec:adaptation}

\begin{figure}[!t]
\centering
\includegraphics[width=0.9\columnwidth]{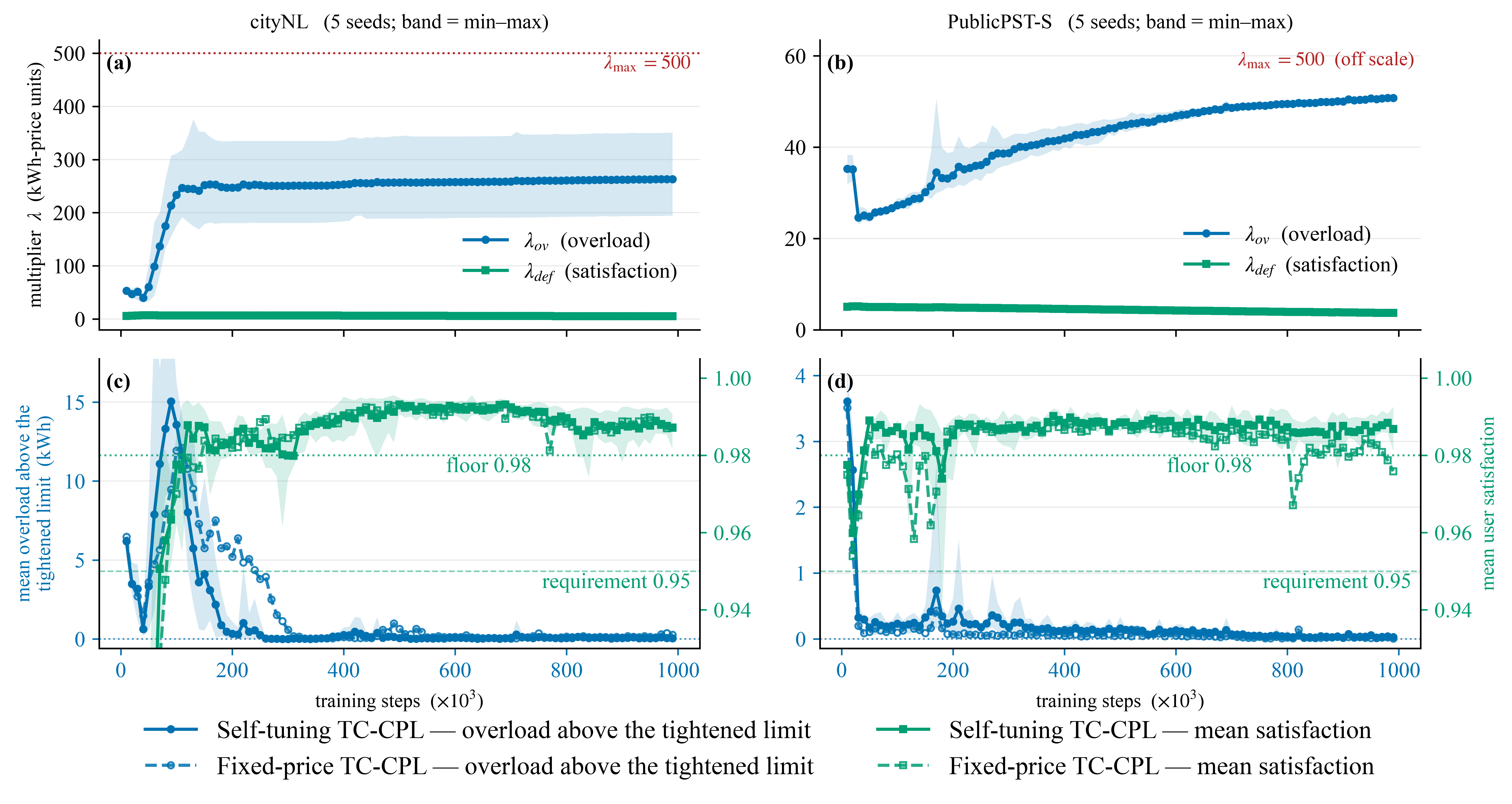}
\caption{Self-tuning TC-CPL on \texttt{cityNL} (left) and \texttt{PublicPST-S} (right). (a), (b)~Lagrange multipliers. (c), (d)~Training overload above the tightened limit and satisfaction per dual window. Seed means with min--max bands (Fixed-price TC-CPL dashed, without bands). Overload nears zero, and $\lambda_{\mathrm{ov}}$ stays below the cap.}
\label{fig:lambda}
\end{figure}

On \texttt{cityNL} (Fig.~\ref{fig:lambda}), $\lambda_{\mathrm{ov}}$ settles early, below the cap, and the overload above the tightened limit nears zero but is rarely exactly zero. Late in training, satisfaction also meets its floor in nearly all windows, so both constraints approximately hold. On \texttt{PublicPST-S}, $\lambda_{\mathrm{ov}}$ ends far lower and still creeps up. The multiplier $\lambda_{\mathrm{def}}$ rises while satisfaction is below its floor and decays toward zero above it, slowly because the slack per window is small.

With $\tau^{\mathrm{ov}}=0$, the settled $\lambda_{\mathrm{ov}}$ is a price at which the constraint holds, not a unique shadow price (Remark~\ref{rem:slater}(i)), and its seed spread reflects early overload, not different outcomes. Fixed-price TC-CPL reaches the same \texttt{cityNL} outcome at a lower price, and neither variant learns clearly faster. The safety gain thus comes from pricing an attainable constraint high enough, by hand or, without a sweep, by adaptation.

With $\rho=0.25$ instead of $1$ in~\eqref{eq:projection}, the rising $\lambda_{\mathrm{ov}}$ signals an apparently unattainable target, not a binding constraint (Remark~\ref{rem:slater}(ii)). On \texttt{cityNL}, even $\delta=-1$ on every port left overload above the tightened limit, no training window came near zero, and the integral grew toward the cap (note to Table~\ref{tab:ablation}). On \texttt{PublicPST-S}, $\lambda_{\mathrm{ov}}$ still rises toward the cap at the end of training, and overload stays well above zero.

\begin{figure}[!t]
\centering
\includegraphics[width=0.9\columnwidth]{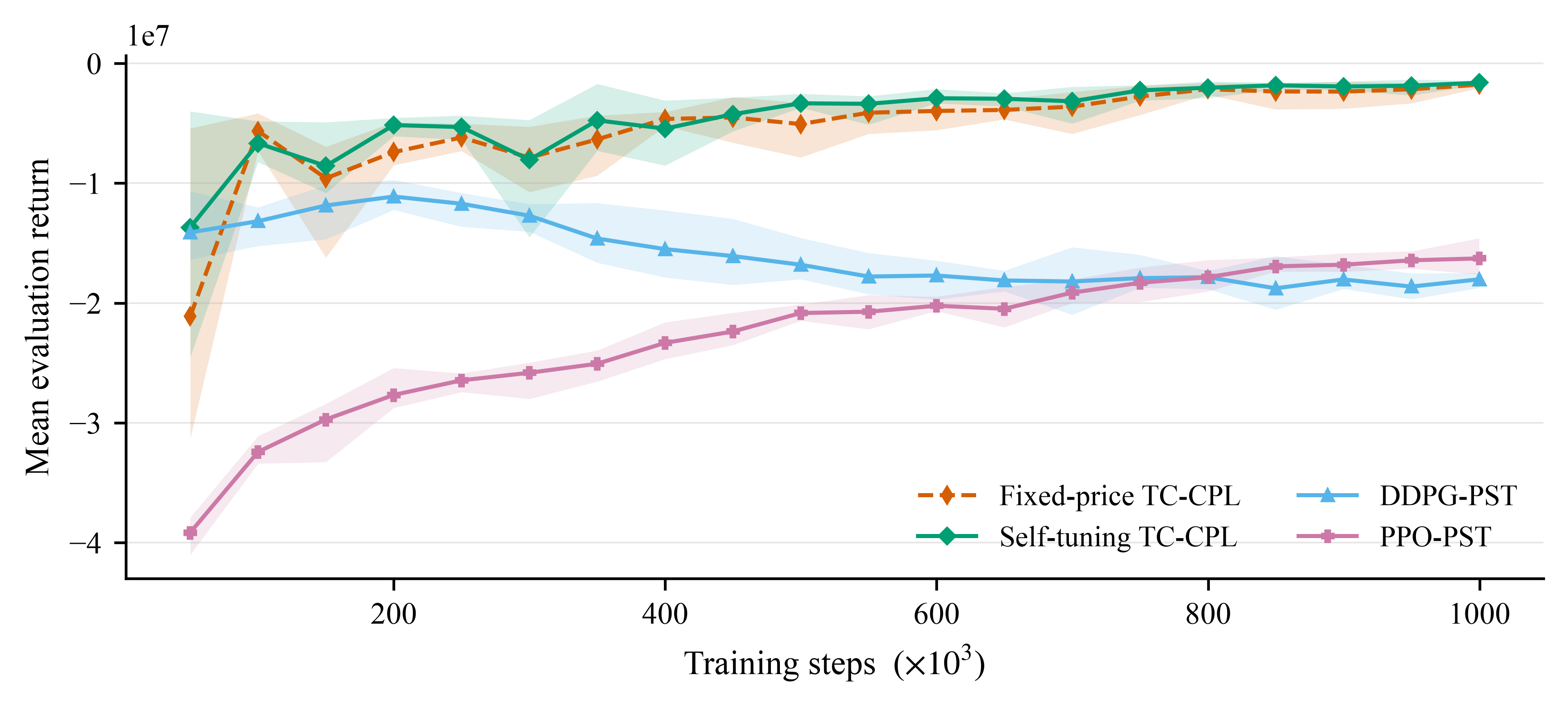}
\caption{\texttt{cityNL} evaluation return (daily sum of the tracking reward~\eqref{eq:stagecost} in kW$^{2}$, without constraint costs) on five simulator-drawn days every $5\times10^{4}$ steps, with deterministic policies (seed means, min--max bands). CPO logs no such return. The TC-CPL variants overlap and end about ten times closer to zero than DDPG-PST and PPO-PST.}
\label{fig:curves}
\end{figure}

M2 does not make training safe (Section~\ref{ssec:m2}). In its first dual window, the learner overloads in both scenarios, on \texttt{PublicPST-S} more than Round-Robin does. The return in Fig.~\ref{fig:curves} ignores overload. DDPG-PST, for example, has the highest return on \texttt{PublicPST-S} (not shown) while overloading often.

\begin{table}[!t]
\centering
\caption{Ablations, sensitivity and fixed prices on \texttt{PublicPST-S}.}
\label{tab:ablation}\label{tab:fixedgrid}
\scriptsize
\newcommand{\abst}{\rlap{$^{*}$}}
\setlength{\tabcolsep}{3.1pt}
\begin{tabular}{@{}lcrrrrrr@{}}
\toprule
Variant & $n$ & $|\epsilon_{\mathrm{tr}}|$ & $\epsilon^{\mathrm{usr}}$ & $E^{\mathrm{del}}$ & $\epsilon^{\mathrm{ov}}$ & OV & $\lambda_{\mathrm{ov}}$ \\
 & & kWh/d & & kWh/d & kWh/d & days\,\% & final \\
\midrule
Self-tuning TC-CPL & 3 & 123.6 & 0.970 & 114.7 & 0.003 & 0.7 & 54 \\
Round-Robin (incumbent) & -- & 100.0\abst & 0.974 & 122.7\abst & 1.858\abst & 49.0\abst & -- \\
\midrule
\multicolumn{8}{@{}l}{\emph{M1 off, prices frozen at $(\lambda_{\mathrm{ov}},\lambda_{\mathrm{def}})$}} \\
$(10,5)$ & 2 & 116.9\abst & 0.978\abst & 119.9\abst & 0.011 & 4.0 & 10 \\
$(10,50)$ & 2 & 116.5\abst & 0.974\abst & 118.3\abst & 1.051 & 27.0 & 10 \\
$(100,5)$ & 2 & 127.2 & 0.955 & 107.6 & 0.010 & 3.0 & 100 \\
$(100,50)$, Fixed-price & 3 & 127.9\abst & 0.961 & 109.6\abst & 0.000 & 0.0 & 100 \\
$(500,5)$ & 3 & 133.3\abst & 0.948\abst & 103.0\abst & 0.000 & 0.0 & 500 \\
$(500,50)$ & 2 & 133.4\abst & 0.942\abst & 101.1\abst & 0.000 & 0.0 & 500 \\
$(2000,50)$ & 2 & 146.6\abst & 0.922\abst & 92.1\abst & 0.000 & 0.5 & 2000 \\
\midrule
\multicolumn{8}{@{}l}{\emph{M2 (warm start and residual)}} \\
No demonstrations & 3 & 125.5 & 0.958\abst & 110.0\abst & 0.001 & 1.0 & 66 \\
No residual anchoring & 3 & 120.4 & 0.982\abst & 119.9\abst & 0.001 & 0.3 & 48 \\
$\rho=0.25$ & 3 & 106.6\abst & 0.977\abst & 122.1\abst & 1.070\abst & 36.7\abst & 342$^{\ddagger}$ \\
CAFAP anchor & 3 & 130.9\abst & 0.984\abst & 120.6\abst & 0.024 & 2.0 & 88 \\
CAFAP anchor, $\rho=0.25$ & 2 & 137.4\abst & 0.998\abst & 130.8\abst & 3.855\abst & 55.0\abst & 500$^{\dagger}$ \\
\midrule
\multicolumn{8}{@{}l}{\emph{M3 (observation and encoder)}} \\
No transformer block & 3 & 173.2\abst & 0.834\abst & 59.0\abst & 0.040\abst & 4.7\abst & 95 \\
Flat MLP encoder & 3 & 129.1\abst & 0.938\abst & 105.3\abst & 1.399\abst & 38.3\abst & 240$^{\ddagger}$ \\
Not load-aware & 3 & 121.3 & 0.979\abst & 119.2\abst & 0.003 & 1.3 & 51 \\
\midrule
\multicolumn{8}{@{}l}{\emph{M1 settings (default $K_p=5$, $K_i=2$, $\tau^{\mathrm{ov}}=0$, $m=0.05$)}} \\
$K_p=0$ (dual ascent) & 2 & 125.5 & 0.975\abst & 117.1\abst & 0.011 & 2.5 & 57 \\
$K_i=0.5$ & 2 & 123.0 & 0.975\abst & 117.9\abst & 0.000 & 0.0 & 24 \\
$K_i=8$ & 2 & 127.9 & 0.964 & 112.0 & 0.000 & 0.5 & 146 \\
$K_p=K_i=2$ & 2 & 122.2 & 0.972 & 117.2\abst & 0.003 & 1.0 & 55 \\
$\tau^{\mathrm{ov}}=0.5$~kWh/day & 2 & 114.0\abst & 0.979\abst & 120.8\abst & 0.032 & 5.0 & 9$^{\ddagger}$ \\
$\tau^{\mathrm{ov}}=2$~kWh/day & 3 & 135.6\abst & 0.945\abst & 102.7\abst & 0.003 & 0.3 & 17$^{\ddagger}$ \\
$m=0$ (no back-off) & 2 & 121.3\abst & 0.977 & 119.2\abst & 0.015 & 3.0 & 48 \\
\bottomrule
\end{tabular}

\smallskip
\begin{minipage}{\columnwidth}
\scriptsize Grouped rows change one element of row~1 ($2\times10^{5}$ steps, $F=3{,}072$, $n$ seeds), and $\lambda_{\max}\in\{100,2000\}$ reproduces row~1 seed for seed. $^{*}$Resolved vs.\ row~1. $^{\dagger}$At the cap. $^{\ddagger}$Unsettled. On \texttt{cityNL}, $\rho=0.25$ leaves $4.24$~kWh/day above the tightened limit even with $\delta=-1$ on every port, and at least $2.4$~kWh/day in every training window.
\end{minipage}
\end{table}

\subsection{Ablations and Sensitivity}
\label{ssec:ablation}

Table~\ref{tab:ablation} changes one element of Self-tuning TC-CPL at a time, in shorter, less converged runs, so close rankings may change. Apart from the $\rho=0.25$ runs of Section~\ref{ssec:adaptation}, no \texttt{cityNL} ablation was run. The shaping term~\eqref{eq:3-hinge} was not ablated.

\textbf{M1 and fixed prices.} Frozen prices trade service against overload, and Fixed-price TC-CPL errs toward safety at a cost in delivered energy. Held fixed, the initial pair $(10,5)$ even serves and tracks better than Self-tuning TC-CPL at an unresolved overload difference, but the right pair is scenario-dependent and takes a sweep to find.

\textbf{M1 settings.} The cap is inactive here, so changing it has no effect (note to Table~\ref{tab:ablation}). It was not varied on \texttt{cityNL}, where $\lambda_{\mathrm{ov}}$ is higher. All gain settings, including the augmented-Lagrangian form ($K_p=K_i=2$)~\cite{chen2025augmented}, settle at a price set by $K_i$, and most deliver slightly more energy than the default at unresolved overload changes. A tolerance $\tau^{\mathrm{ov}}>0$ makes the integrator two-sided, so $\lambda_{\mathrm{ov}}$ cycles down to zero. With a small tolerance, tracking and service improve, and with a large one, service drops. Without the back-off ($m=0$), tracking and delivered energy improve at an unresolved overload change, so learning, not the back-off, achieves most of the reduction from Round-Robin.

\textbf{M2.} With CAFAP as the incumbent, $\rho=0.25$ drives $\lambda_{\mathrm{ov}}$ to the cap, and at $\rho=1$, CAFAP trades tracking for service. Residual anchoring does not help here. Removing it ($a_t=(\delta_t+1)/2$, no $a^0_t$ input, demonstrations kept) raises satisfaction with no resolved overload change, as does removing both M2 elements (two seeds). As the demonstrations help only with the anchoring, M2 shows no measured benefit at $\rho=1$.

\textbf{M3.} The transformer block~\eqref{eq:obs-w} is essential, and a flat MLP encoder worsens every metric, including overload. The load-aware inputs do not help, as removing them raises satisfaction at unchanged overload, possibly (untested) because $P^{\mathrm{pot}}_t$ in~\eqref{eq:obs-g} already reveals the number of connected vehicles at a single transformer.

\subsection{Deployment Checks}
\label{ssec:robust}

\subsubsection{\textbf{Headroom projection}}
\predisplaypenalty=0
An optional dispatch layer, used in no other result, enforces the transformer limit without retraining. With $\tilde a_{j,t}=a_{j,t}/\max\{1,\sum_{j'\in\mathcal{J}_i}a_{j',t}\}$, the command of port $j\in\mathcal{J}_i$ normalized over all ports of its station (occupied or not), each port $j\in\mathcal{J}_w$ receives
\begin{equation}
\label{eq:7-shield}
\begin{aligned}
a^{\mathrm{s}}_{j,t}&=\tilde a_{j,t}\min\bigl\{1,\,H^{m_{\mathrm{s}}}_{w,t}/\widehat U_{w,t}\bigr\},\\
H^{m_{\mathrm{s}}}_{w,t}&=\pos{(1-m_{\mathrm{s}})\overline{P}^{\mathrm{tr}}_{w,t}-P^{\mathrm{nEV}}_{w,t}},\\
\widehat U_{w,t}&=\textstyle\sum_{i\in\mathcal{C}_w}\overline{P}^{\mathrm{cs}}_{i}\min\bigl\{1,\sum_{j'\in\mathcal{J}_i}a_{j',t}\bigr\},
\end{aligned}
\end{equation}
with margin $m_{\mathrm{s}}\ge 0$. Because ports draw at most their commands, $P^{\mathrm{tr}}_{w,t}\le\max\{(1-m_{\mathrm{s}})\overline{P}^{\mathrm{tr}}_{w,t},P^{\mathrm{nEV}}_{w,t}\}$ holds in every interval, grid side included, up to the nowcast error of the non-EV load. On the ablation models, the projection removed all overload of Self-tuning TC-CPL and Round-Robin, at negligible energy cost for TC-CPL (Table~\ref{tab:robust}(a)). Projected Round-Robin still tracks better than TC-CPL in Table~\ref{tab:results} but serves less.

\subsubsection{\textbf{Observation errors}}
The tests in Table~\ref{tab:robust}(a) corrupt only the learner's observation. Port-status errors hurt most, as a misread port receives the limit-blind Round-Robin command, yet overload stays an order of magnitude below Round-Robin's. Non-EV load and rating errors matter much less, and removing the DR look-ahead leaves overload unchanged. Unlike on a real site, the incumbent saw the true status, and \texttt{cityNL} was not tested.

\begin{table}[!t]
\centering
\caption{Deployment checks on \texttt{PublicPST-S} and computational cost.}
\label{tab:robust}\label{tab:compute}
\scriptsize
\def\hd#1{\begin{tabular}[b]{@{}>{\bfseries\boldmath}c@{}}#1\end{tabular}}%
\setlength{\tabcolsep}{0pt}
\begin{tabular*}{\columnwidth}{@{\extracolsep{\fill}}lrrrrrr@{}}
\toprule
\textbf{(a) Deployment checks} & \multicolumn{4}{c}{\bfseries Self-tuning TC-CPL} & \multicolumn{2}{c@{}}{\bfseries Fixed-price} \\
\cmidrule(lr){2-5}\cmidrule(l){6-7}
\textbf{Observation error} & \hd{$\Delta\epsilon^{\mathrm{ov}}$} & \hd{Seed-\\days} & \hd{Max\\excess} & \hd{$\Delta\epsilon^{\mathrm{usr}}$} & \hd{$\Delta\epsilon^{\mathrm{ov}}$} & \hd{Seed-\\days} \\
($10^{6}$ steps, five seeds) & kWh/d & \% & kW & & kWh/d & \% \\
\midrule
None (absolute values) & 0.0014 & 0.6 & 0.48 & 0.985 & 0.0004 & 0.4 \\
\addlinespace[2pt]
Status flipped, 10\% & $+0.111$ & 13.4 & 8.46 & $+0.002$ & $+0.106$ & 12.8 \\
Status hidden, 10\% & $+0.149$ & 13.8 & 9.64 & $+0.002$ & $+0.129$ & 13.2 \\
Port data 15~min late & $+0.163$ & 22.6 & 7.15 & $+0.002$ & $+0.134$ & 22.0 \\
\addlinespace[2pt]
Non-EV load $\times1.2$ & $-0.001$ & 0.0 & 0.00 & $-0.009$ & $+0.001$ & 0.4 \\
Non-EV load $\times0.8$ & $+0.011$ & 4.4 & 1.65 & $+0.006$ & $+0.010$ & 3.6 \\
Non-EV load of $t{-}1$ & $+0.019$ & 6.4 & 3.40 & $+0.001$ & $+0.013$ & 3.4 \\
\addlinespace[2pt]
Rating $\times0.95$ & $+0.004$ & 0.4 & 2.12 & $-0.005$ & $0.000$ & 0.0 \\
Rating $\times1.05$ & $+0.003$ & 1.2 & 0.96 & $+0.004$ & $+0.003$ & 1.2 \\
No DR notice & $0.000$ & 0.6 & 0.49 & $+0.001$ & $0.000$ & 0.4 \\
\cmidrule{1-7}
\multicolumn{7}{@{}l}{\rlap{\emph{Headroom projection}~\eqref{eq:7-shield}, $m_{\mathrm s}=0$, $2\times10^{5}$-step models, three seeds}} \\
\textbf{Controller} & \multicolumn{3}{c}{\bfseries Overload} & \hd{$E^{\mathrm{del}}$ cost} & \hd{$|\epsilon_{\mathrm{tr}}|$} & \hd{$\epsilon^{\mathrm{usr}}$} \\
 & kWh/d & \% & kW & \% & kWh/d & \\
\cmidrule(l){2-7}
Self-tuning TC-CPL & 0 & 0.0 & 0.00 & 0.035 & -- & -- \\
Round-Robin & 0 & 0.0 & 0.00 & 1.9 & 102.3 & 0.971 \\
\end{tabular*}\par\nointerlineskip
\begin{tabular*}{\columnwidth}{@{\extracolsep{\fill}}lrr@{}}
\midrule
\textbf{(b) Computational cost} & \hd{\texttt{PublicPST-S}} & \hd{\texttt{cityNL}} \\
\midrule
Whole controller, mean / p99 (ms) & 0.15 / 0.35 & 2.31 / 3.10 \\
TC-CPL training (h), mean [min, max] & 15.8 [10.0, 20.8] & 22.6 [12.3, 28.7] \\
DDPG-PST / PPO-PST / CPO (h) & 2.4 / 0.30 / 0.34 & 8.1 / 5.2 / 5.8 \\
SAC update, laptop CPU / GPU (ms) & 23.7 / 14.3 & 746 / 237 \\
Memory, update / replay (GB) & 0.05 / 0.17 & 1.50 / 5.12 \\
DDPG-PST / PPO-PST / CPO (k param.) & 37 / 52 / 651 & 483 / 573 / 2{,}509 \\
\end{tabular*}\par\nointerlineskip
\begin{tabular*}{\columnwidth}{@{\extracolsep{\fill}}lrrrrr@{}}
\cmidrule{1-6}
\textbf{TC-CPL parameters} & \hd{$\chi_{\mathrm c}$} & \hd{$\chi_{\mathrm w}$} & \hd{$\chi_{\mathrm g}$} & \hd{Head} & \hd{Total} \\
\cmidrule(l){2-6}
Actor ($0.47$~MB) & 17{,}152 & 33{,}664 & 17{,}152 & 49{,}666 & 117{,}634 \\
Each critic & 17{,}280 & 33{,}664 & 17{,}152 & 33{,}025 & 101{,}121 \\
\end{tabular*}\par\nointerlineskip
\begin{tabular*}{\columnwidth}{@{\extracolsep{\fill}}lrrrrr@{}}
\cmidrule{1-6}
\multicolumn{6}{@{}l}{\rlap{\emph{Synthetic layouts}, $|\mathcal{W}|=N/5$ (trained \texttt{cityNL} actor, random observations)}} \\
\hd{$N$} & \hd{MFLOP} & \hd{Actor + proj.} & \hd{Headroom proj.} & \hd{Activations} & \hd{$|\mathcal{W}|{\times}N$ matrix} \\
 & (pooling \%) & ms & ms & MB & MB \\
\cmidrule(l){2-6}
20 & 3.0 (1) & 0.11 & 0.02 & 0.08 & ${<}0.01$ \\
600 & 106 (18) & 0.46 & 0.45 & 2.3 & 0.29 \\
20{,}000 & 23{,}725 (88) & 38.0 & 25.9 & 76 & 320 \\
\bottomrule
\end{tabular*}

\vspace{2pt}
\parbox{\columnwidth}{\scriptsize $\Delta$ is the change from the unperturbed run. A flipped status swaps charging and full, and a hidden one shows the port empty, each with probability $0.1$ per port and interval. Latencies (batch~1, one CPU thread of a shared Apple M5 Pro laptop) are upper bounds.}
\end{table}

\subsection{Computational Cost and Scalability}
\label{ssec:compute}

\leavevmode TC-CPL's parameter count is site-independent (Section~\ref{ssec:m3}), unlike the baselines', though DDPG-PST and PPO-PST are smaller on \texttt{PublicPST-S} (Table~\ref{tab:compute}(b)). On \texttt{cityNL}, the whole controller takes five orders of magnitude less time than the dispatch interval, even on a shared laptop, and on the largest synthetic layout the actor and projection still take only tens of milliseconds. Training dominates the cost, with SAC updates over ten times costlier on \texttt{cityNL} and times that include contention on shared GPU servers.

\section{Conclusion}
\label{sec:conclusion}

This paper formulated transformer-constrained PST as a CMDP using only the measurements an operator has, without departure times or states of charge, and proposed TC-CPL, which extends soft actor-critic with three mechanisms. A PI controller sets the overload and satisfaction multipliers from violations measured in training (M1). The policy learns a residual on the site's incumbent dispatcher, starting from that dispatcher's demonstrations (M2). A transformer-aware encoder shared by all ports keeps the parameter count independent of the site size (M3).

In both scenarios, TC-CPL cut the incumbent Round-Robin's transformer overload by orders of magnitude, to near zero, and raised user satisfaction. On the district, it also tracked better and delivered more energy. On the single stressed transformer, it tracked worse than Round-Robin, which kept most of that lead even when projected to zero overload, so tracking is TC-CPL's shortfall there. CAFAP, DDPG-PST and PPO-PST overloaded more and tracked worse than TC-CPL, and CPO nearly avoided overload only by barely charging. Against Fixed-price TC-CPL, the M1 ablation with hand-set multipliers, self-tuning was statistically indistinguishable on the district and served better on the single transformer. A fixed pair with lower multipliers served as well or better, but other pairs in the sweep lost service or overloaded, so a fixed choice must be searched for in each scenario, whereas the adaptation needed one setting for both. With the full residual radius, the overload multiplier stayed below its cap; with a small one, the target was apparently unattainable and the multiplier kept rising.

\textit{Limitations.} All results come from simulation. The transformer model has no thermal dynamics or aging (Section~\ref{ssec:statement}). The overload constraint is met only approximately, and~\eqref{eq:7-shield} enforces it per interval only if the non-EV load is known. Except in Section~\ref{ssec:robust}, port status and non-EV load were observed exactly. The \texttt{cityNL} DR schedule repeats daily. The ablations, limited to one facility with few seeds, showed no measured benefit from M2 or the load-aware inputs. Dense-pooling memory grows with $|\mathcal{W}|N$. Training needs the departure state of charge and overloads transformers while exploring (Section~\ref{ssec:m2}), so it belongs in simulation. We would not use TC-CPL without reliable per-interval port-status, session-meter and non-EV load data, without the projection where small overloads are unacceptable, or without retraining when a site's layout or load changes.

\section*{Code and data availability}
The implementation is openly available at \url{https://github.com/danialsj77/TC-CPL}.
 
\bibliographystyle{IEEEtran}
\bibliography{Ref}

\end{document}